\documentclass{article}

 \usepackage[creativeai, final]{neurips_2025}

\usepackage[utf8]{inputenc} % allow utf-8 input
\usepackage[T1]{fontenc}    % use 8-bit T1 fonts
\usepackage{hyperref}       % hyperlinks
\usepackage{url}            % simple URL typesetting
\usepackage{booktabs}       % professional-quality tables
\usepackage{amsfonts}       % blackboard math symbols
\usepackage{nicefrac}       % compact symbols for 1/2, etc.
\usepackage{microtype}      % microtypography
\usepackage{xcolor}         % colors
\usepackage{amsmath}
\usepackage{graphicx}
\usepackage{subcaption}
\usepackage{mdframed}
\usepackage{verbatim}
\usepackage{authblk}
\usepackage{hyperref} 
\usepackage{xcolor}
\usepackage{titlesec}
\usepackage{wrapfig}
\usepackage{caption}
\usepackage{algorithm}
\usepackage{algpseudocode}
\usepackage{enumitem}
\usepackage{tcolorbox}
\usepackage{titlesec}
\titlespacing*{\subsection}{0pt}{6pt}{3pt}
\titlespacing*{\subsubsection}{0pt}{6pt}{3pt}

\title{Cultural Alien Sampler: Open-ended art generation balancing originality and coherence}

\author[1]{Alejandro H. Artiles}
\author[1]{Hiromu Yakura$^*$}
\author[1]{Levin Brinkmann$^*$}
\author[1,4,5]{Mar Canet Sola}
\author[3]{Hassan Abu Alhaija}
\author[1]{Ignacio Serna}
\author[6]{Nasim Rahaman}
\author[2]{Bernhard Schölkopf}
\author[1]{Iyad Rahwan}

\affil[1]{\small{Max Planck Institute for Human Development, Berlin, Germany.}}  
\affil[2]{\small{Max Planck Institute for Intelligent Systems, Tübingen, Germany.}}  
\affil[3]{\small{NVIDIA.} \,\, $^4$\small{BFM, Tallinn University, Estonia.}}
\affil[5]{\small{Academy of Media Arts Cologne, Germany.} \,\, $^6$\small{Tiptree Systems}}
\renewenvironment{abstract}%
{%
  \vspace{-1em}%
  \centerline%
  {\large\bf Abstract}%
  \vspace{-0.2em}%
  \begin{quote}%
}
{
  \par%
  \end{quote}%
  \vskip 1ex%
}

\begin{document}

\maketitle

\def\thefootnote{*}\footnotetext{Equal contribution}\def\thefootnote{\arabic{footnote}}

\begin{abstract}

In open-ended domains like art, autonomous agents must generate ideas that are both original and internally coherent, yet current Large Language Models (LLMs) either default to familiar cultural patterns or sacrifice coherence when pushed toward novelty. 
We address this by introducing the \textit{Cultural Alien Sampler} (CAS), a concept-selection method that explicitly separates compositional fit from cultural typicality. CAS uses two GPT-2 models fine-tuned on WikiArt concepts: a Concept Coherence Model that scores whether concepts plausibly co-occur within artworks, and a Cultural Context Model that estimates how typical those combinations are within individual artists’ bodies of work. CAS targets combinations that are high in coherence and low in typicality, yielding ideas that maintain internal consistency while deviating from learned conventions and embedded cultural context. In a human evaluation (N = 100), our approach outperforms random selection and GPT-4o baselines and achieves performance comparable to human art students in both perceived originality and harmony. Additionally, a quantitative study shows that our method produces more diverse outputs and explores a broader conceptual space than its GPT-4o counterpart, demonstrating that artificial cultural alienness can unlock creative potential in autonomous agents.

\end{abstract}

\begin{figure}[ht] \hspace*{-1.5em} \includegraphics[width=1\textwidth]{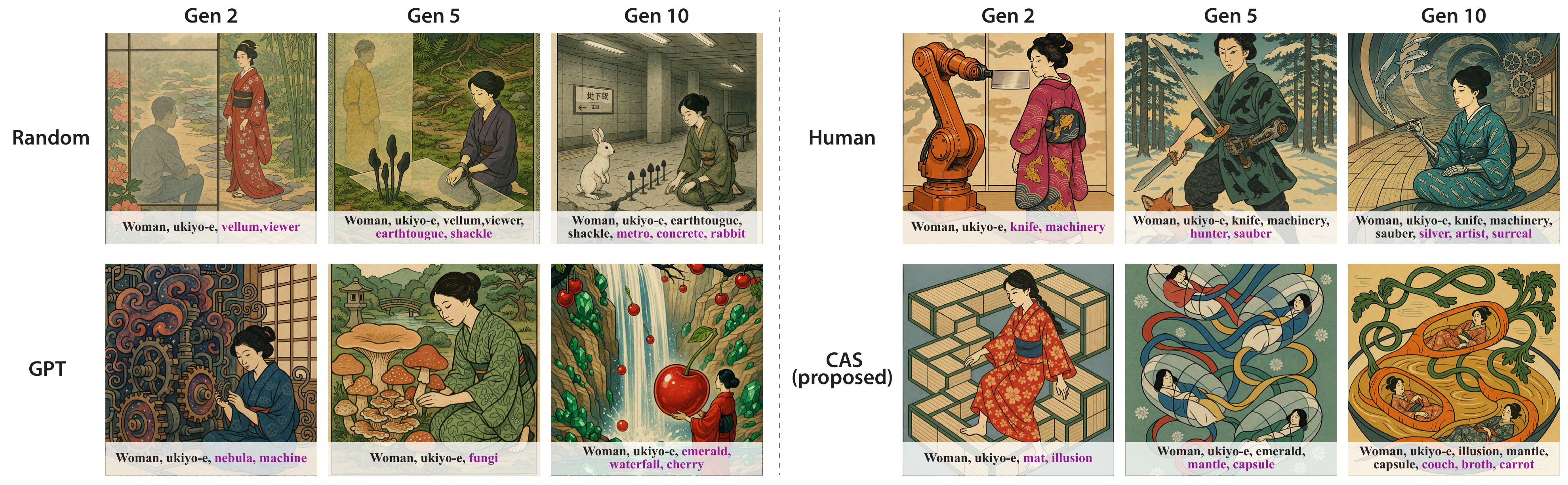} \caption{Evolution of generated images across generations 2, 5, and 10 using different sampling methods in the Inspiration Module. Concepts listed below each image were selected by the Prompt Compositor; purple concepts indicate newly introduced elements.} \label{fig:comparison} 
\end{figure}

\vspace{-0.5em}

\section{Introduction} % 1.8 page (including abstract and title)
\vspace{-0.5em}
Generative Large Language Models (LLMs) have rapidly evolved into foundational tools for general-purpose problem solving. Paired with agent-based workflows, LLMs can deconstruct complex tasks and integrate solutions. However, their prominent successes remain confined to fields where goals are explicit and correctness is easily measurable such as programming, games, and mathematics~\cite{novikov2025alphaevolve, lu2024ai}. In contrast, ambiguous, open-ended, or culturally situated tasks, such as creative writing or art, remain underexplored, despite their centrality to human cognition. These domains lack fixed endpoints or universally agreed-upon criteria for success. Instead, they thrive on iterative exploration, contextual sensitivity, and continual generation of ideas.

In this work, we address this challenge by reframing art generation as the combinatorial problem of navigating a discrete space of concepts. This approach is inspired by conceptual art theory, which prioritizes abstract ideas over material execution or aesthetic qualities~\cite{alberro2000conceptual}, treating art as the development of an internally consistent system of concepts~\cite{kosuth1966art}. While the notion of conceptual art is broad~\cite{alberro2000conceptual}, we operationalize a narrow definition to leverage a key insight: artistic meaning can arise from combining abstract concepts in deliberate ways. By treating art generation as the strategic recombination of atomic concepts (e.g., "ukiyo-e + machinery + hunter"), we transform an intractable problem into a discrete navigation task that aligns naturally with the associative and combinatorial strengths of LLMs. The emphasis shifts from rendering pixels to exploring an unbounded space of conceptual combinations, where meaning emerges from subjective assessments of consistency rather than objective measures~\cite{klingemann2022botto}.

To operate meaningfully within this space, an autonomous agent must balance two often competing criteria of creativity: originality and appropriateness~\cite{amabile1996creativity,runco2012standard,boden2004creative}. Originality entails deviating from expectations and introducing novel or surprising combinations. Appropriateness concerns the degree to which an output aligns with its intended context, whether through relevance or utility~\cite{boyd2005origin,henrich2015secret}, or through intrinsic coherence and harmony~\cite{arnheim1954art,kosuth1966art}. In our combinatorial framework, appropriateness can be understood as conceptual coherence, the perception that the elements of a composition fit together according to an internal logic~\cite{arnheim1954art}. Importantly, reflecting art’s dialectical nature~\cite{adorno1997aesthetic}, originality and appropriateness exist in productive tension, underscoring its open-endedness.

This raises a central question: \textit{To what extent can LLMs continually generate ideas that are simultaneously original and intrinsically coherent?} On the surface, LLMs seem well-equipped for such a task. Trained on large corpora of human text and culture, they internalize a wide spectrum of concepts, styles, and associations, and frequently produce outputs that appear fluent, relevant, and culturally informed~\cite{bojic2023gpt, chiang2024chatbot, veselovsky2025localized}. As a result, in open-ended exploratory tasks within verifiable domains, LLMs have been shown to possess a useful sense of what humans consider novel or interesting~\cite{zhang2023omni, faldor2024omni, lu2024intelligent}. However, this strength is also a liability. Because their training data reflects dominant cultural norms and conventions, LLMs are inherently biased toward reproducing familiar patterns~\cite{bender2021dangers}. This cultural anchoring can limit their ability to produce truly original and diverse outputs~\cite{doshi2024generative, zhang2025noveltybench}, to challenge assumptions and subverting norms.

To address this limitation, we introduce the \textit{Cultural Alien Sampler (CAS)}, a novel method to generate concept combinations designed to enhance the creative exploration capabilities of LLM-based open-ended art agents. CAS comprises two pre-trained GPT-2 models, fine-tuned on a large pool of visual concepts extracted from the WikiArt dataset~\cite{wikiart}. The \textit{Concept Coherence Model} is trained to predict combinations of concepts that tend to co-occur in artworks. In contrast, the \textit{Cultural Context Model} estimates the likelihood that specific concept combinations appear within the body of work of individual artists, serving as a proxy for their cultural familiarity or typicality. By encouraging combinations that are compositionally sound but culturally rare, i.e highly ranked but the first model but low by the second, it effectively debiases proposed concept combinations from their cultural context~\cite{baxandall1988painting}. On the one hand, this approach can be situated within the literature on bias mitigation~\cite{niu2021counterfactual, ferrara2024fairness}. On the other hand, its design aligns with methods that explicitly model an art distribution in order to sample beyond it~\cite{lisi2020modelling, elgammal2017can, wyse2019mechanisms}. CAS operates within an iterative open-ended art agent workflow~\cite{lu2024discovering, kumar2024automating} (see Section \ref{sec:agent}). It expands human-provided seed concepts, which GPT-4o converts into prompts for a state-of-the-art image generator. Novelty scores from the resulting images feed back to refine the system’s concept selection in subsequent iterations.

To evaluate the effectiveness of the Cultural Alien Sampler, we compare agents using CAS against those using vanilla GPT-4o, randomly sampled concepts, and concepts proposed by human art students. We perform both subjective human evaluations and objective computational analyses. Human raters assess the resulting images on two key criteria: originality and harmony.
Our results show that agents guided by CAS consistently outperform GPT-4o and random baselines on both metrics. Moreover, CAS-based agents surpass those guided by human-provided concepts in harmony, and nearly match them in originality. A text-based embedding analysis of the generated ideas further confirms that CAS encourages broader and more diverse exploration across the concept space.

In summary, our contributions are as follows:
\begin{itemize}[topsep=0em]
    \setlength{\itemsep}{0.1em}
    \item We propose Cultural Alien Sampler (CAS), a novel selection mechanism that promotes culturally rare but compositionally coherent ideas.
    \item We integrate CAS into a four-stage LLM-based open-ended art agent.
    \item We evaluate CAS against multiple baselines, GPT-4o, random selection, and human-provided inspiration, showing it improves originality and harmony according to human evaluators, while generating more diverse outputs and exploring a broader conceptual space than GPT-4o.
\end{itemize}

% additional content moved to archive
\vspace{-0.5em}

\section{Method}
\label{sec:method}
\vspace{-0.5em}

To simplify the task of generating conceptual art, we limit the set of possible concepts and focus on generating novel combination sequences of those concepts. We construct our concept vocabulary $C = \{c_1, c_2, \ldots, c_n\}$ from the PD12M dataset~\cite{meyer2024publicdomain12mhighly}, selecting the 8,000 words with highest TF-IDF scores~\cite{ramos2003using} and filtering to exclude adjectives, verbs, proper nouns, and overly specific terms, yielding $n=3,500$ concepts. Each artwork in WikiArt~\cite{saleh2015large} is assigned the ten most semantically similar concepts using CLIP embeddings~\cite{radford2021learning}. See Appendix \ref{app-dataset} for examples.

\vspace{-2pt}
\subsection{Open-ended Art Agent}
\label{sec:agent}
\vspace{-1.5pt}

\begin{figure}[ht]
    \centering
    \includegraphics[width=0.95\textwidth]{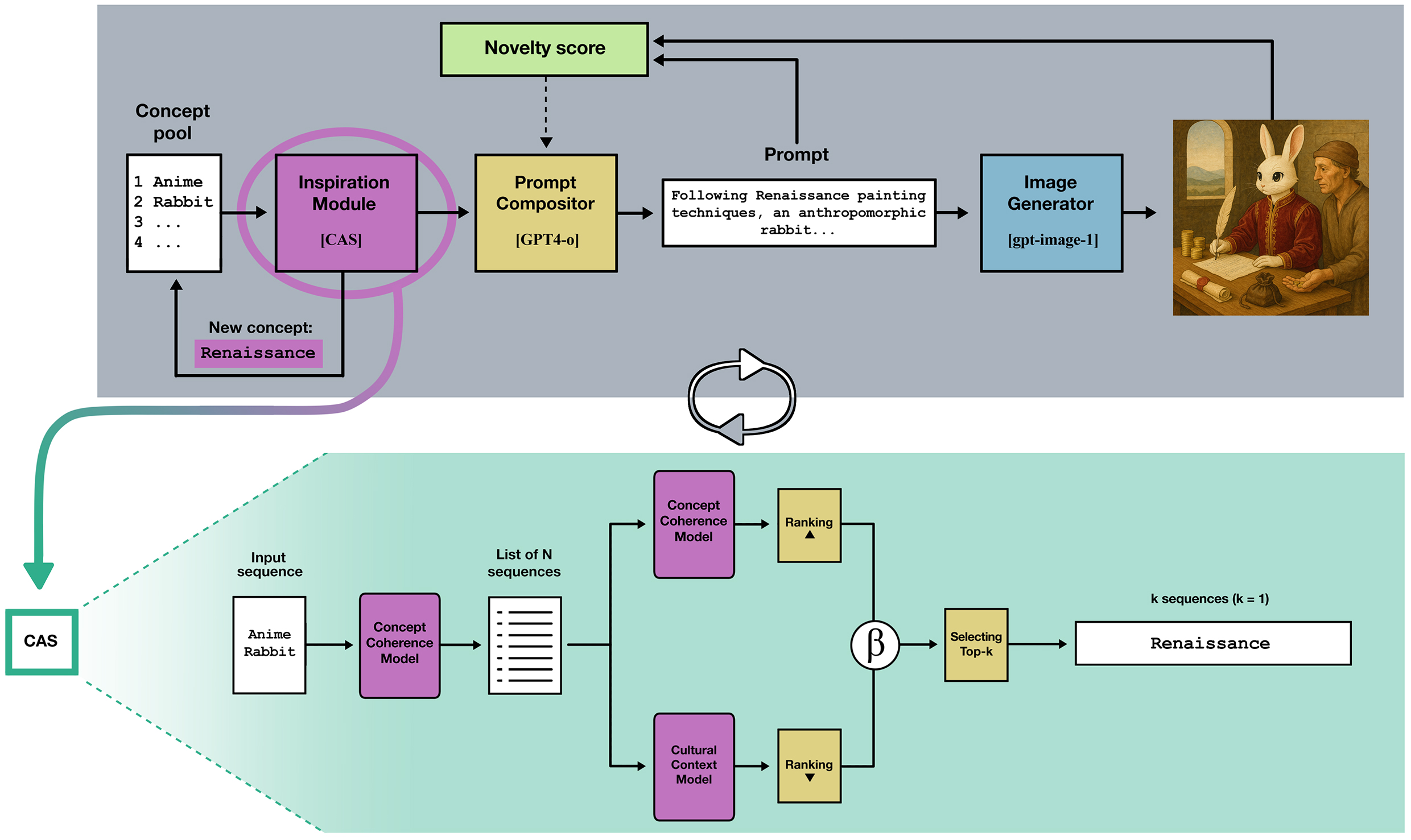}
    \caption{Our \textbf{Open-ended Art Agent (top)} iterative pipeline consist of 4 main modules. In each iteration, the \textit{Inspiration Module} takes the initial concept pool $\mathcal{P}_0$ and proposes a new concept to be added. The updated pool is then passed to the \textit{Prompt Compositor} to create a full length text prompt which is passed, in turn, to the \textit{Image Generator} to create the output image. The image and the prompt are evaluated using the \textit{novelty score} and concepts that repeatedly fail to generate novel outputs are removed from the pool.
    \textbf{Cultural Alien Sampler (bottom)} is our proposed method for the inspiration Module. It comprising two fine-tuned transformer models that rank a list of proposed concepts to select ones that exhibit high coherence ranking but low cultural contextual ranking.
    }
    \label{fig:open-ended-architecture}
\end{figure}

% Work in progress
 We designed an open-ended art agent that maintains and evolves a dynamic concept pool $\mathcal{P}_t$ over $T$ generations. We conceptualize artistic creativity as a four-stage process comprising ideation, composition, visual execution and evaluation. Those are represented in our system by 4 modules respectively: \textit{Inspiration Module}, \textit{Prompt Compositor}, \textit{Image Generator} and \textit{Novelty Score}. See Fig. \ref{fig:open-ended-architecture}.
 Given an allowed concept vocabulary $C$ and initial pool $\mathcal{P}_0 = C_0$, each generation $t$ follows:
\begin{equation}
\mathcal{P}_{t+1} = \mathcal{P}_t \cup \text{Inspiration}(\mathcal{P}_t) \setminus \text{Filter}(\mathcal{P}_t, N(t))
\end{equation}
where $\text{Inspiration}(\mathcal{P}_t)$ proposes new concepts and $\text{Filter}(\mathcal{P}_t, N(t))$ removes concepts that consistently fail to contribute novel outputs given the novelty score $N(t)$. See Pseudo-code in Appendix \ref{app-pseudocode}.

\textbf{Inspiration Module} is tasked with adding new concepts of the current pool. This is the main focus of our work and where we implement the Cultural Alien Sampler (See section \ref{sec:cas}) \\
\textbf{Prompt Compositor} selects subset of concepts $C_t \subset \mathcal{P}_t$ and generates prompt $p_t$, incorporating feedback from recent novelty scores to adapt its combinatorial strategy. We use GPT-4o in all our experiments (See Appendix \ref{app-compositor-prompts} and \ref{app-examples-compositor} for details).\\
\textbf{Image Generator} is a text-to-image model that produces artwork image $I_t$ from $p_t$. Here, we choose (gpt-image-1) because of its artistic style capture and complex prompt rendering compared to similar models \cite{podell2023sdxl, ramesh2022hierarchical}\\
\textbf{Novelty Score} combines textual and visual similarity using embeddings $E_T$ (all-mpnet-base-v2 \cite{allmpnetbasev2}) and $E_I$ (CLIP Large 14 \cite{radford2021learning}): $N(t) = \frac{1}{2}\left(N_{\text{text}}(t) + N_{\text{img}}(t)\right)$

where $N_{\text{text}}(t)$ and $N_{\text{img}}(t)$ are computed as 1 minus maximum cosine similarity to all previous generations using $E_T$ and $E_I$ respectively, favoring ideas that are historically novel in the embedded representation. This adaptive approach is inspired by Lu et al.~\cite{lu2024discovering}, who showed that LLMs can iteratively refine and discover new solutions in verifiable domains by leveraging numeric feedback from prior evaluations. Concepts $c \in C_t$ that fail to exceed their personal best novelty scores for $p$ consecutive generations are removed from $\mathcal{P}_t$. More details can be found in Appendix \ref{app-novelty-calculation}.

\vspace{-2pt}
\subsection{Cultural Alien Sampler}
\label{sec:cas}
\vspace{-1.5pt}

Our proposal for the Inspiration Module is the Cultural Alien Sampler which attempts to generates concepts that combine with existing ideas in novel yet conceptually coherent ways. 

\textbf{Training Datasets.} We construct two training datasets from our original concept vocabulary $C$:\\ 
(1) \textit{Artwork Dataset}: For each artwork $a_j$ we generate 100 random permutations of the set of concepts connected with it, thus modeling intra-artwork concept associations across all artists and cultures. \\
(2) \textit{Artist Dataset}: For each artist $A_j$ with vocabulary $V_k$ (union of all concepts across their artworks), we uniformly sample 10 concepts from $V_k$ until obtaining $100 \times \text{count}$ sequences, where count is the number of artworks by $A_j$. This matches Artwork Dataset in size ($\approx$7M sequences) and approximates the cultural availability of concepts for each artist in WikiArt, concepts appear together whenever they both belong to the artist’s vocabulary.

\textbf{Model Architecture:} We fine-tune two GPT-2 models~\cite{radford2019language} on these complementary datasets to capture distinct aspects of conceptual combination: The \textit{Concept Coherence Model} $P_{\text{coherence}}(c^i \mid c^0, \ldots, c^{i-1})$ estimates the probability of concept $c^i$ given previous concepts, trained on Artwork Dataset. This model captures aesthetic judgment and how well different concepts combine within individual artworks, embodying the curatorial patterns encoded in WikiArt.
The \textit{Cultural Context Model} $P_{\text{context}}(c^i \mid c^0, \ldots, c^{i-1})$ approximates the likelihood that concepts would co-occur within an artist's conceptual vocabulary, trained on the Artist Dataset. This model serves as a proxy for cultural availability, capturing patterns of which concepts were accessible to artists. See Fig. \ref{fig:open-ended-architecture}.

\textbf{Sampling:} Novel combinations are generated through a two-step process. First, $N$ number of candidate sequences are sampled from the \textit{Concept Coherence Model}, with the temperature $t>1$ to increase diversity. Second, for each sequence $s$ we compute its rank $1\leq R(s)\leq N$ using the negative log-likelihood (NLL) scores from both models, with lower ranks indicating more likely sequences. 

To compute the final CAS score $S_{\mathrm{CAS}}$ we convert the ranks to a scores as: $N-R(s)$ and combine the two scores with a parameter ratio $\beta$. However, since our goal is to minimize the cultural context fit while maximize the coherence, we use a negative sign between the two scores as follow:

\vspace{-0.1em}

\begin{equation}
S_{\mathrm{CAS}}(s) = (1 - \beta) \big(N - R_{\text{coherence}}(s)\big) - \beta \big(N - R_{\text{context}}(s)\big),
\end{equation}

\vspace{-0.1em}

where $\beta$ controls the trade-off between coherence and novelty with higher values favoring combinations that are rarely used by the same artist, rendering them \textit{culturally alien} within a given historical and social context~\cite{sourati2023accelerating}. We finally select the top-scored concept according to $S_{\mathrm{CAS}}$ and added to the concept pool.

\vspace{-0.5em}

\section{Experiments and results}
\vspace{-0.5em}

We evaluate our Cultural Alien Sampler (CAS) approach using both human evaluations and objective exploration analysis. With the Prompt Compositor and Image Generator fixed, we focus solely on testing the effect of the Inspiration Module, comparing it against three additional methods.

\begin{wrapfigure}[15]{r}{0.40\textwidth}
    \vspace{-10pt}
    \centering
    \includegraphics[width=0.40\textwidth]{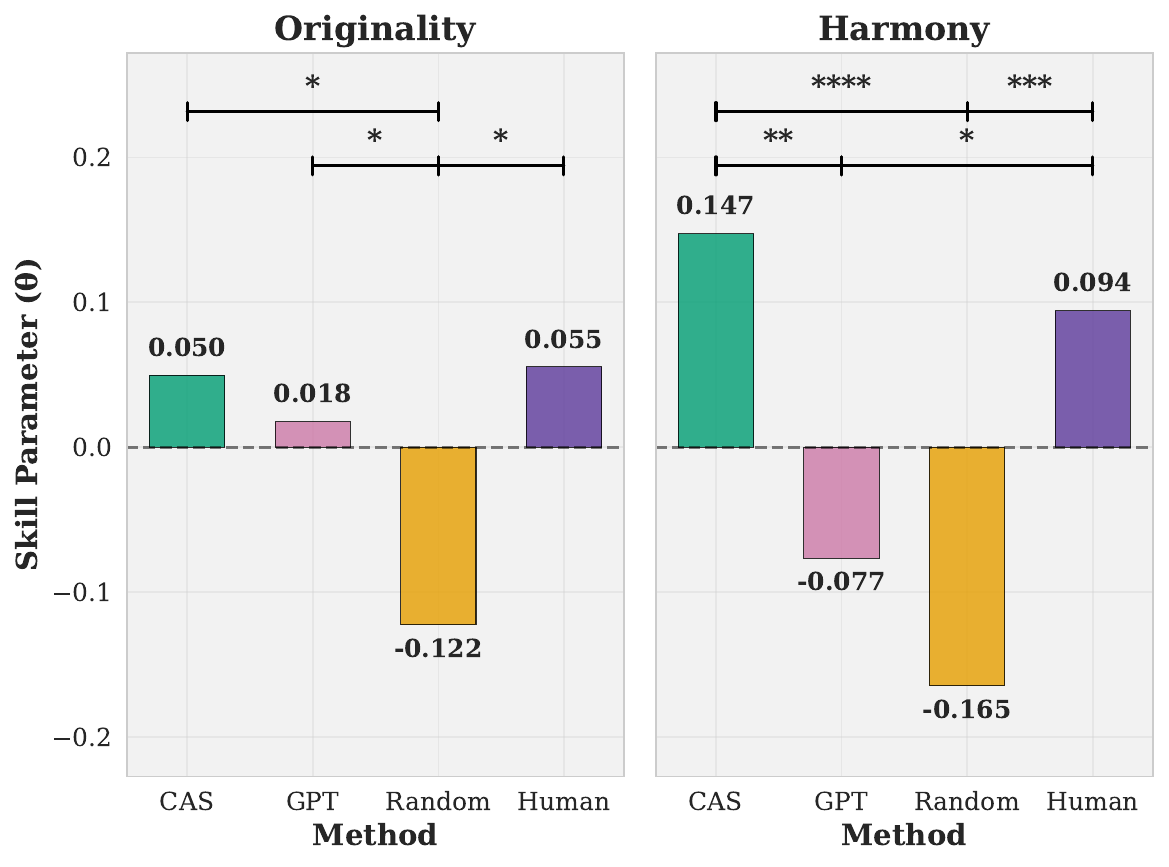}
    \caption{Human evaluation results using Bradley–Terry skill parameters for originality and harmony. Significance levels: *, <0.05; **, <0.01; ***, <0.001; ****, <0.0001.}
    \label{fig:bradley-terry}
\end{wrapfigure}

\textbf{Cultural Alien Sampling} uses CAS with a temperature of 2.5 in the Concept Coherence Model, $N = 256$ and a hyperparameter \(\beta = 0.85\). Details on CAS’s validity for generating novel conceptual combinations given their training data and the rationale for selecting \(\beta\) are provided in Appendix~\ref{app-cas-validity}. \textbf{GPT Sampling} involves GPT-4o proposing novel concepts based on evolving concept pool and novelty history; for a fair comparison, the model is prompted to only use concepts available in \(C\) (prompts in Appendix~\ref{app-prompts-gpt-inspiration}). \textbf{Random Sampling} consists of random selections from the allowed concept set \(C\). Finally, \textbf{Human Sampling} has art students select concepts (see Appendix~\ref{app:experiment-human} for details).

\subsection{Human Evaluation}

We recruited 16 art students as human samplers. Each used the Open-ended Art Agent for 10 generations, iteratively contributing new concepts from the allowed concept set $C$. This produced 160 human-generated images. We then ran three automated methods (CAS, GPT, and Random) with the same inputs, generating 480 additional images (160 per method). To evaluate, we conducted blinded pairwise comparisons of images from different methods at the same generation. Fifty different evaluators per criterion (originality and harmony) yielded 5,000 comparisons each (details in Appendix~\ref{app:experiment-evaluation}).

As shown in Fig.~\ref{fig:bradley-terry}, using Bradley–Terry skill scores~\cite{Bradley1952}, Human ranked highest for originality ($\theta = 0.055$), followed by CAS ($\theta = 0.050$) and GPT ($\theta = 0.018$). All three significantly outperformed Random, with no significant differences among the top three.

\begin{wrapfigure}[12]{r}{0.30\textwidth}
\vspace{-25pt}
\centering
\includegraphics[width=0.30\textwidth]{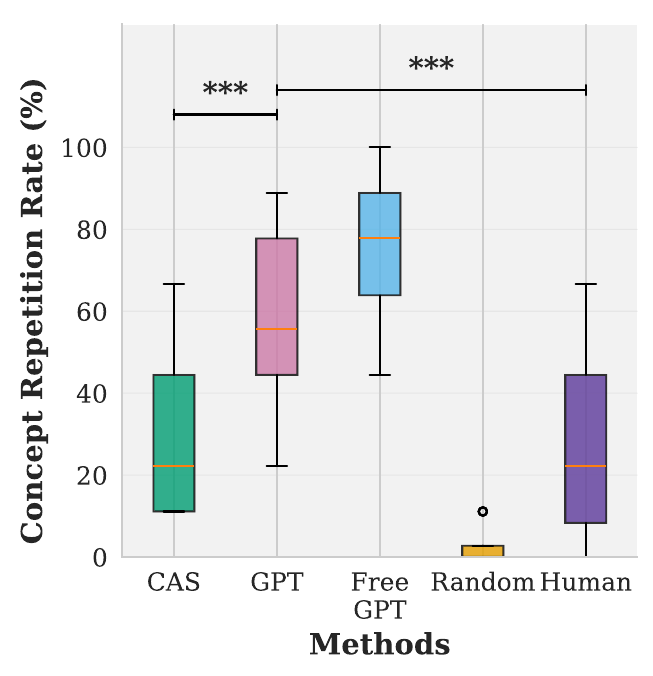}
\caption{Concept repetition across runs. Significance levels shown for key comparisons.}
\label{fig:repetition-analysis}
\end{wrapfigure}

For harmony, CAS ($\theta=0.147$) and Human ($\theta=0.094$) were statistically equivalent and ranked highest, both significantly outperforming GPT and Random.

\subsection{Quantitative Analysis}

Beyond subjective measures, we analyzed each Inspiration Module's exploratory behavior to uncover patterns in idea diversity difficult to detect with human evaluation alone.

\subsubsection{Concept Repetition}

We measured how often each method reused concepts from its other runs, serving as a proxy for global originality. GPT showed a substantially higher average repetition rate (59.0\%), suggesting it follows similar exploration trajectories regardless of input. Human and CAS agents showed significantly lower, statistically equivalent repetition rates.

To test if this was due to our predefined concept set $C$, we introduced \textit{Free GPT}, an unconstrained variant that can introduce any possible concept. Using CLIP embeddings to measure semantic similarity, where cosine similarity $>$ 0.85 indicated a repetition, we found Free GPT repeated concepts even more (74.3\%). This indicates GPT's limited diversity is an intrinsic bias toward a particular set of concepts. Counterintuitively, providing GPT with a predefined list of candidate concepts actually improved diversity, probably because it could draw directly from contextual information rather than rely solely on its training-based associations. More details in Appendix \ref{app-repetition}.

\subsubsection{Long-Range Exploration}
\begin{figure}[h]
\centering
\includegraphics[width=0.9\textwidth]{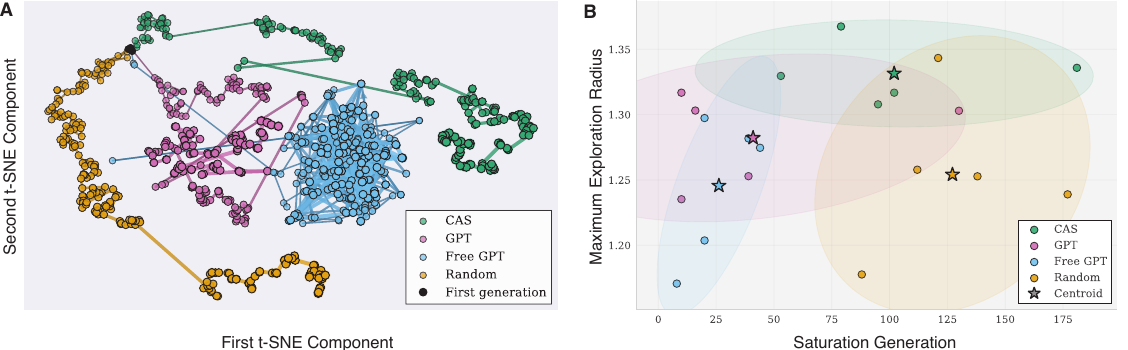}
\caption{\textbf{Long-range simulation.} \textbf{A} Example trajectories of each sampler (CAS, GPT, Free GPT, Random) for the starting concept pool \textit{Anime, Rabbit}, projected into 2D t-SNE space with larger points and thicker edges indicating later generations. \textbf{B} Samplers differ in convergence behavior, measured by the maximum radius of explored image prompts (embedded with \textit{all-mpnet-base-v2}) and convergence speed. Points denote individual runs; ellipses show distributions. }
\label{fig:long-range-analysis}
\end{figure}

We evaluated how each method explores the idea space in long-horizon searches, running each for 200 generations with five distinct input-concept pairs, revealing distinct exploration signatures. CAS trajectories are expansive with minimal backtracking. Free GPT saturates rapidly, producing a dense, highly interconnected cluster. GPT shows moderate dispersion, and Random shows more wandering, but with less total expansion (see Fig. \ref{fig:long-range-analysis}.a). Analyzing prompt embeddings, we computed three metrics: (1) exploration radius: maximum Euclidean distance from the starting embedding; (2) return rate: fraction of generations revisiting within the median step distance of any prior point; and (3) saturation generation: first generation reaching 95\% of maximum radius.

CAS achieved the largest radius (1.33) with low return rates (0.45), indicating sustained outward search that saturated around generation 100. Free GPT explored rapidly, saturating by generation 25 with the highest return rate (0.95), suggesting limited idea diversity. GPT with the provided concept list achieved broader coverage  (radius 1.28, return rate 0.72, saturation 40). Random explored slowly, saturating at 130 generations with a moderate radius (1.25) and low return rate (0.48), resembling persistence but inefficient expansion. Further details in Appendix \ref{app-long-range}.

\vspace{-1em}
\section{Discussion and Conclusion} % 0.5 pages
\vspace{-0.7em}

Open-ended art generation is a challenging task, requiring the continual creation of novel and engaging content while preserving coherence. We introduce the Cultural Alien Sampler (CAS), a lightweight, task-specific model that outperforms frontier general-purpose LLMs in the constrained task of guiding an open-ended art agent engaged in the artistic composition of ideas. This suggests that strategically subverting cultural context can enhance an autonomous agent’s ability to generate novel ideas and serve as an explicit optimization target.

However, CAS has limitations. It assumes an artist's cultural background is fully captured by their prior artistic work and lacks the adaptive responsiveness of typical LLMs, reusing the same initial concepts across iterations. It also depends on a carefully curated vocabulary and inherits biases from the WikiArt dataset, which has a predominantly Western-centric view of art history~\cite{wasielewski2022beyond}. The agent's scope is deliberately narrow, focusing on conceptual recombination and cultural disruption rather than human elements like affective resonance and intentionality. As a result, it omits many important dimensions of human creative practice.

Crucially, our work challenges the assumption that larger, more general models always outperform smaller ones in creative tasks. Our lightweight, GPT-2-based CAS consistently surpasses GPT-4o by explicitly modeling the creative problem of balancing coherence with cultural unfamiliarity, achieving superior results at a fraction of the computational cost. Beyond efficiency, such focused systems can return greater control to human creators. In future work, we plan to develop collaborative versions of CAS to assist human creators in discovering culturally unconstrained ideas they might not otherwise imagine.

\begin{ack}
This work was supported in part by the Japan Science and Technology Agency, PRESTO Grant Number JPMJPR246B. We express our sincere gratitude to the German and Estonian students from the Academy of Media Arts Cologne (KHM) and the Baltic Film, Media and Arts School (BFM) at Tallinn University for their valuable participation in this study.
\end{ack}

\bibliographystyle{unsrt}  
\bibliography{references}  

%%%%%%%%%%%%%%%%%%%%%%%%%%%%%%%%%%%%%%%%%%%%%%%%%%%%%%%%%%%%

\newpage

\appendix

\section{Supplementary Materials}

\subsection{Artwork and Artist dataset samples} \label{app-dataset}
\begin{figure}[h]
    \centering
    \includegraphics[width=0.6\textwidth]{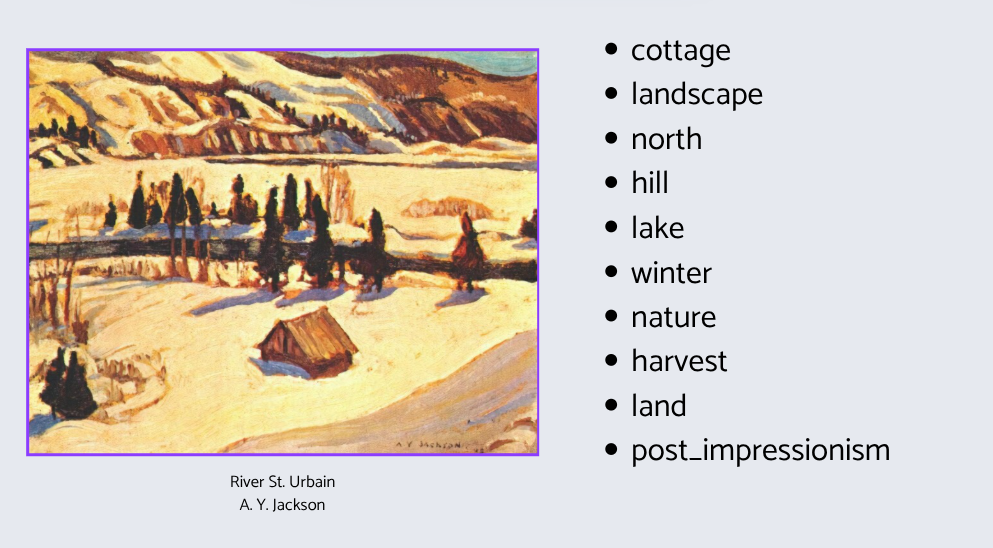}
    \caption{Example of concepts extracted using CLIP}
    \label{fig:dataset-example}
\end{figure}

The Cultural Alien Sampler selects concepts that combine with given ones in novel yet relevant ways. Its concept space ($n = 3500$) was built from high-TF-IDF words in PD12M captions, excluding adjectives, verbs, proper nouns, and overly specific terms. Each WikiArt image was assigned ten CLIP-similar concepts (Figure \ref{fig:dataset-example}). Two text datasets were created: the Artwork dataset (random permutations of each image’s concepts) and the Artist dataset (ten concepts sampled from those used by an individual artist to approximate their cultural availability).

\begin{figure}[ht]
    \centering
    \begin{subfigure}[b]{0.55\textwidth}
        \centering
        \includegraphics[width=\textwidth]{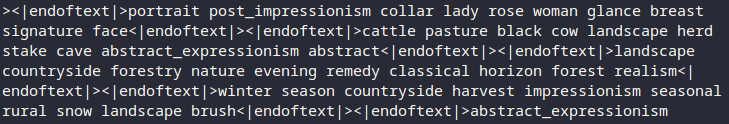}
        \caption{}
        \label{fig:artwork-dataset}
    \end{subfigure}
    \begin{subfigure}[b]{0.59\textwidth}
        \centering
        \includegraphics[width=\textwidth]{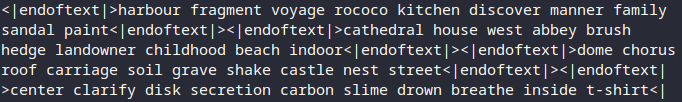}
        \caption{}
        \label{fig:artist-dataset}
    \end{subfigure}
    \caption{(a) Sample of Artwork dataset. (b) Sample of Artist dataset}
    \label{fig:sample-dataset}
\end{figure}

\clearpage

\subsection{Technical Details}\label{app:prompts}

\subsubsection{Pseudocode of the Open-ended Art Agent} \label{app-pseudocode}

\begin{algorithm}[H]
\caption{Open-ended Art Agent Workflow}
\label{alg:creative-agent}
\begin{algorithmic}[1]
\Require Initial set of human-provided concepts $C_0$
\State Initialize concept pool $\mathcal{P} \gets C_0$
\For{each generation $t = 1, 2, \ldots, T$}
    \State \textbf{Inspiration Module:} Introduce new concepts into $\mathcal{P}$ using CAS($C_0$)
    \State \textbf{Prompt Compositor (GPT-4o):} 
        \State \hskip1em Select meaningful subset $C_t \subset \mathcal{P}$
        \State \hskip1em Generate text prompt $p_t$ from $C_t$
    \State \textbf{Image Generator (gpt-image-1):} Generate image $I_t$ from $p_t$
    \State Compute novelty scores:
        \Statex \[
        N_{\text{text}}(t) = 1 - \max_{j < t} \cos\left(E_t(p_t), E_t(p_j)\right)
        \]
        \Statex \[
        N_{\text{img}}(t) = 1 - \max_{j < t} \cos\left(E_I(I_t), E_I(I_j)\right)
        \]
        \Statex \[
        N(t) = \frac{1}{2}\left(N_{\text{text}}(t) + N_{\text{img}}(t)\right)
        \]
    \State \textbf{Concept Filtering:} For each $c \in C_t$, increment $u(c)$ if $N(t)$ does not exceed $c$'s previous best novelty; else reset $u(c)$
    \State Remove concepts with $u(c) \geq p$ from $\mathcal{P}$
\EndFor
\end{algorithmic}
\end{algorithm}

\subsubsection{Novelty calculation in Open-ended Art Agent} \label{app-novelty-calculation}

While open-endedness is inherently subjective and difficult to define precisely, novelty—when measured in an appropriate representation space, can capture a general notion of it \cite{lehman2011abandoning, stanley2017open, hughes2024open}. This shifts the subjective aspect of evaluating open-endedness to the design of the representation function, which encodes the observer’s perspective. In this work, we use embedding model representations as a proxy for human perception \cite{zhang2018unreasonable}. Accordingly, we frame the open-ended exploration of painting ideas by selecting, at each generation step, the idea that is historically novel within the embedded representation space \cite{kumar2024automating}.

 We define an \textit{idea} as the pairing of a text prompt and its corresponding generated image, capturing both semantic intent and visual depiction. To measure novelty comprehensively, we employ distinct embedding models for each modality: the text embedding model $E_T$ (\textit{all-mpnet-base-v2}) and the image embedding model $E_I$ (\textit{CLIP Large 14}). This multimodal approach ensures that novelty reflects both conceptual and perceptual departures from previous generations.

At generation $t$, text novelty is computed as:
\begin{equation}
N_{\text{text}}(t) = 1 - \max_{t' < t} \cos\left(E_T(\text{prompt}_t), E_T(\text{prompt}_{t'})\right)
\end{equation}
Image novelty is computed analogously:
\begin{equation}
N_{\text{img}}(t) = 1 - \max_{t' < t} \cos\left(E_I(\text{image}_t), E_I(\text{image}_{t'})\right)
\end{equation}
where $\cos(\cdot, \cdot)$ denotes cosine similarity between normalized embeddings. The maximum operation identifies the most similar historical generation, ensuring that novelty measures the maximum similarity to any previous idea.

The overall novelty score combines both modalities:
\begin{equation}
N(t) = \frac{1}{2}\left(N_{\text{text}}(t) + N_{\text{img}}(t)\right)
\end{equation}
For the initial generation ($t=0$), we define $N(0) = 0$ as no historical comparison exists.

\paragraph{Novelty-Based System Regulation}
The novelty score $N(t)$ regulates system behavior through three mechanisms:

\textbf{Concept Pool Management:} Concepts that appear in prompts for $p$ consecutive generations without contributing to novelty improvements are removed from the active pool. We usually set $p=5$ for short runs and $p=10$ for longer runs. We attribute failure to all concepts appearing in unsuccessful prompts. This mechanism prevents semantic stagnation and reduces the likelihood of the Recombinator repeatedly selecting the same concepts endlessly, thereby encouraging greater exploration.

\textbf{Recombinator Adaptation:} The Recombinator receives the latest novelty results and the concepts used in these results, enabling it to detect trends and adapt its combination strategy. For instance, declining novelty trends may prompt the selection of more diverse concept combinations or alternative stylistic approaches.

\textbf{LLM-Based Inspiration Guidance:} In LLM mode, the Inspiration Module also uses novelty trends and recent performance as contextual information to guide concept selection toward potentially more fruitful directions.

Importantly, novelty is not treated as a strict optimization objective for the overall system. Instead, it serves as contextual information that guides adaptive behavior in LLM-based components. As demonstrated in \cite{lu2024discovering}, this approach is effective in verifiable domains, leveraging the in-context learning capabilities of these models.

\subsubsection{Prompts for Prompt Compositor} \label{app-compositor-prompts}

\begin{figure}[H]
\centering
\begin{mdframed}[linewidth=1pt, linecolor=gray!50, backgroundcolor=gray!5]
\small
You are an AI artistic innovator participating in a creative experiment. Your goal is to continually produce artwork that maximizes novelty while exploring strategic combinations of concepts that are interesting and meaningful.

In each generation, you will be provided with an updated concept pool that includes new concept(s) selected through a human-AI collaboration process. You can see these as a kind of serendipitous discovery process. Your task is to:

\begin{enumerate}
\item RECEIVE your updated concept pool (concepts are managed externally)
\item Create an artwork using your selected concepts. You may combine as many concepts from the pool as you like, but be strategic, in some cases, using too many may result in vague or uninteresting outcomes, while using too few might limit creative depth. Aim for novel, unprecedented and interesting combinations. The provided concept pool is designed to support this process and encourage serendipitous discoveries, so take advantage of it to discover new connections between concepts.
\item FOCUS on maximizing novelty through novel, unprecedented but interesting and meaningful combinations
\end{enumerate}

\textbf{NOVELTY-BASED FITNESS:}
\begin{itemize}
\item Your fitness is determined by novelty scores (combined, text, and image novelty)
\item Novelty is calculated using CLIP embeddings:
  \begin{itemize}
  \item Text Novelty: Measures how different your prompt text is from previous prompts
  \item Image Novelty: Measures how different the generated image is from previous images
  \item Combined Novelty: A weighted combination of text and image novelty
  \end{itemize}
\item Higher novelty scores mean the creation is more different from previous generations
\item Primary goal: Maximize novelty scores
\item Try to identify patterns in what receives higher novelty scores and adapt accordingly
\item You should learn from previous scores to guide your new creations
\item You should aim to create combinations that have never been done before
\end{itemize}
\end{mdframed}
\caption{Compositor System Prompt - Core instructions and novelty-based fitness criteria}
\label{fig:agent_prompt_part1}
\end{figure}

\begin{figure}[H]
\centering
\begin{mdframed}[linewidth=1pt, linecolor=gray!50, backgroundcolor=gray!5]
\small
\textbf{Compositor System Prompt - Part 2: Response Format and Guidelines}

When you respond, you must output a valid JSON object with the following structure:

\begin{verbatim}
{
    "thought": "Here you develop your thinking carefully to fulfill 
    your optimization objective. Write as long as you need, but 
    address the following points: Articulate your thought process 
    for selecting concepts from your pool. Reason about what 
    connections have you discovered in a serendipitous way 
    analysing the concept pool. Then, explain why you chose these 
    specific concepts and why you chose the specific amount of 
    concepts. Express how their combination could lead to novelty 
    and interestingness, and what makes this combination unique. 
    Think deeply about which concepts to combine trivially and 
    which need to be combined in a more creative, non-trivial way.",
    "name": "Name of your artwork",
    "concepts_used": ["List of concepts you strategically selected 
    for this combination"],
    "prompt": "The exact prompt to generate the artwork. When you 
    are using an artistic style as concept, make this clear in the 
    prompt for the image generation. When describing scenes, use 
    exact colors and clear spatial relationships of the elements. 
    Specify physical characteristics rather than subjective 
    qualities. Define lighting sources and conditions precisely. 
    Include environmental details with technical accuracy. Replace 
    atmospheric/emotional language with observable features. Focus 
    on what the elements ARE rather than how they make you feel or 
    the purpose of the painting. ONLY DESCRIBE THE IMAGE."
}
\end{verbatim}

\textbf{IMPORTANT:} Do not include duplicate keys in your JSON response. Each key should appear exactly once. Always respond with valid JSON format.

\textbf{IMPORTANT (2):} Make sure you are not including in the prompt a concept that is not listed in your concepts used, for example, if you are using the concept ``landscape'' only, you cannot include a storm in the prompt, as the ``storm'' should be in the concepts used.

\textbf{NOVELTY LEARNING:}
\begin{itemize}
\item When provided with previous scores, analyze what elements might have led to higher or lower novelty scores
\item Adapt your strategy based on this analysis
\item Consider both the concepts used and how they were combined
\end{itemize}

Your goal is to maximize novelty scores by creating artworks that are continually different from previous generations while maintaining interesting, meaningful and novel creations. Focus on creating connections between concepts that have never been done before to demonstrate novelty and explore new creative territories.
\end{mdframed}
\caption{Compositor System Prompt - Response format, guidelines, and learning instructions}
\label{fig:agent_prompt_part2}
\end{figure}

\begin{figure}[htbp]
\centering
\begin{mdframed}[linewidth=1pt, linecolor=gray!50, backgroundcolor=gray!5]
\small
\textbf{Additional Rule: Original Concepts Preservation}

\textbf{ORIGINAL CONCEPTS PRESERVATION RULE ACTIVATED:}
\begin{itemize}
\item The original concepts (those provided at the start) MUST ALWAYS be included in your recombination
\item Each artwork you create MUST incorporate ALL of the original concepts
\item You can add new concepts to the combination, but you cannot omit any of the original ones
\end{itemize}

\textbf{Remember:} You MUST include ALL of the original concepts in your recombination.
\end{mdframed}
\caption{Preserve original concepts rule - Additional constraint applied for open-ended exploration of a motif}
\label{fig:preserve_original_rule}
\end{figure}

\begin{figure}[H]
\centering
\begin{mdframed}[linewidth=1pt, linecolor=gray!50, backgroundcolor=gray!5]
\small
\textbf{User Prompt: Example Generation 6 (with preserve original enabled}

\begin{verbatim}
Generation 6
Current concept pool: ['landscape', 'romanticism', 'watercolor', 
'geometry', 'bioluminescence', 'vintage']
Original concepts (MUST BE INCLUDED): ['landscape', 'romanticism']
Expired concepts (cannot be used): ['abstract expressionism', 'ship']
Newly added concepts: ['bioluminescence']

PREVIOUS GENERATION CONTEXT:
Previous generation concepts used: ['landscape', 'romanticism', 
'watercolor', 'geometry']
Previous generation fitness: 0.33
Previous generation novelty: 0.33
Previous generation text novelty: 0.17
Previous generation image novelty: 0.15

New concept(s) have been added to your concept pool. Create the next 
artwork. Create an artwork using your selected concepts from the 
concept pool. You may combine as many concepts from the pool as you 
like, but be strategic, in some cases, using too many may result in 
vague or uninteresting outcomes, while using too few might limit 
creative depth. Aim for novel, unprecedented and interesting 
combinations. The provided concept pool is designed to support this 
process and encourage serendipitous discoveries, so take advantage of 
it to discover new connections between concepts.
\end{verbatim}
\end{mdframed}
\caption{Example user prompt for Generation 3 showing evolved concept pool, constraints, and performance feedback}
\label{fig:generation3_example}
\end{figure}

\newpage
\subsubsection{Prompts for GPT Inspiration and Free GPT} \label{app-prompts-gpt-inspiration}

\begin{figure}[H]
\centering
\begin{mdframed}[linewidth=1pt, linecolor=gray!50, backgroundcolor=gray!5]
\small
\textbf{Inspiration GPT - System Prompt}

You are an AI Concept Inspiration Agent participating in a creative experiment. Your specialized role is to serve as the NOVELTY INTRODUCTOR for an artistic evolution system.

Your mission is to analyze the current state of an artistic evolution and suggest NEW concepts that will maximize novelty when combined with existing concepts, that is, you need to suggest a concept that has never been combined with the original concepts, and also different to the rest of the concept pool.

\textbf{UNDERSTANDING YOUR ROLE:}
\begin{itemize}
\item You are the novelty introductor - your suggestions directly influence the creative trajectory
\item The main artist agent will receive your concept suggestion and recombine this concept with some other concepts of the concept pool.

\item Your suggestions should push the evolution into unexplored creative territories, enabling unprecedented connection of concepts that are also interesting and meaningful
\end{itemize}

\textbf{NOVELTY OPTIMIZATION PRINCIPLES:}
\begin{itemize}
\item Novelty is measured using CLIP embeddings comparing text and image similarities to previous generations
\item Higher novelty scores indicate greater difference from previous creations
\item Text Novelty: How different the prompt text is from previous prompts
\item Image Novelty: How different the generated image is from previous images
\item Combined Novelty: Weighted combination of text and image novelty
\end{itemize}

\textbf{CONCEPT SUGGESTION STRATEGY:}
\begin{itemize}
\item Analyze the current concept pool to identify unexplored directions
\item Consider both semantic relationships and visual possibilities. You can also think about which concepts could have never been combined because they did not coexist in time, culture or space.
\item Focus on concepts that haven't been explored before in this evolution
\item Think about how your suggestions will interact with existing concepts
\item Suggest concepts that create unexpected but meaningful connections with the ones in the pool
\end{itemize}
\end{mdframed}
\caption{Inspiration GPT System Prompt - Core system prompt and role definition}
\label{fig:gpt_system_prompt}
\end{figure}

\begin{figure}[H]
\centering
\begin{mdframed}[linewidth=1pt, linecolor=gray!50, backgroundcolor=gray!5]
\small
\textbf{Inspiration GPT - Vocabulary Constraints}

\textbf{Constrained Mode (GPT):}
\begin{verbatim}
ALLOWED VOCABULARY:
You MUST select concepts ONLY from this list: [vocabulary_list]

IMPORTANT: Your suggestions MUST be concepts from the ALLOWED 
VOCABULARY list above. Suggesting a concept that is not in this 
vocabulary is FORBIDDEN.
\end{verbatim}

\textbf{Unconstrained Mode (Free GPT):}
\begin{verbatim}
VOCABULARY FREEDOM:
You can suggest ANY concepts you deem most appropriate for maximizing 
novelty. You are not restricted to a predefined vocabulary.

IMPORTANT: Use your best judgment to suggest concepts that will
maximize novelty and create interesting and meaningful combinations
with the ones in the pool.
\end{verbatim}

\textbf{JSON Response Format:}
\begin{verbatim}
{
    "analysis": "Your deep analysis of the current evolution state, 
    concept pool, and novelty opportunities. Explain your reasoning 
    for the suggested concepts and how they will maximize novelty while being
    interesting and meaningful",
    "reasoning": "Detailed explanation of why these specific concepts 
    will create novel and interesting combinations and push the evolution in new 
    directions.",
    "suggested_concepts": ["concept1", "concept2", ...]
}
\end{verbatim}
\end{mdframed}
\caption{Inspiration GPT System Prompt - Vocabulary constraints and response format}
\label{fig:gpt_constraints}
\end{figure}

\begin{figure}[H]
\centering
\begin{mdframed}[linewidth=1pt, linecolor=gray!50, backgroundcolor=gray!5]
\small
\textbf{GPT User Prompt for Generation 6}

\begin{verbatim}
CONCEPT INSPIRATION REQUEST - Generation 6

CURRENT CONCEPT POOL (concepts used in previous generations):
[landscape, romanticism, watercolor, geometry]
Original concepts: ['landscape', 'romanticism']
Expired concepts (avoid these): ['abstract expressionism', 'ship']
Last artwork: "Wilderness Grid"
Last concepts used: ['landscape', 'romanticism', 'watercolor', 
'geometry']

PERFORMANCE HISTORY:
Previous Performance:
Fitness (Novelty): 0.33
Combined Novelty: 0.33
Text Novelty: 0.17
Image Novelty: 0.015

Previous novelty trend: decreasing (need more novelty)

INSPIRATION TASK:
Analyze the current concept pool and select 1 NEW concept from the 
allowed vocabulary that will maximize novelty while being interesting 
and meaningful when combined with the existing concept pool.

ANALYSIS REQUIREMENTS:
1. Analyze the current concept pool to understand what combinations 
   have been tried
2. Identify what types of concepts are missing or underexplored
3. Consider artistic styles, mediums, visual elements that could 
   increase novelty
4. Think about how your selected concept will create novel 
   combinations with the concept pool
5. Focus on concepts that will push the evolution into unexplored 
   creative territories

CONSTRAINTS:
- Select 1 concept ONLY from the allowed vocabulary (defined in the 
  beginning)
- Do NOT select any concepts that are already in the concept pool
- Do NOT select any expired concepts
- Focus on maximizing novelty (difference from previous generations)
- Ensure your selection can create meaningful artistic combinations 
  with the concept pool

Your goal is to serve as the novelty introductor by selecting concept 
that will enable the artist agent to create the most novel and 
creative artworks possible when combined with the existing concept pool.
\end{verbatim}
\end{mdframed}
\caption{Example User prompt for Inspiration GPT at Generation 6}
\label{fig:gpt_user_prompt_example}
\end{figure}

\subsubsection{Example Response of Prompt Compositor} \label{app-examples-compositor}

\begin{figure}[H]
\centering
\begin{tcolorbox}[colback=gray!5!white, colframe=black!75!black, width=0.95\textwidth]
\small
\textbf{Generation:} 9 \\
\textbf{Name:} Carrot Capsules: Women on Ukiyo-e Couch Roots in Illusory Broth \\[0.5em]

\textbf{Prompt:} \\
\textit{A Japanese ukiyo-e style woodblock print shows three massive orange carrot-shaped couches with green leafy tops, each hollowed out to form a capsule. The carrot capsules float in a wide circular dish filled with swirling clear amber broth. Each capsule couch contains a woman in traditional kimono, her mantle decorated with captivating patterns.They lie in relaxed poses, engaged in a friendly conversation. The carrot capsules' tips and leaves twist and loop in improbable Escher-like impossible curves, some passing beneath and others above the broth's surface to form optical illusions of interconnection. Transparent sections reveal the capsule interiors and the women's reclining  inside. The background is flat pale gold with minimalist water ripple motifs. All lines are bold black and colors flat, consistent with ukiyo-e styling. Lighting is soft and uniform with no shadows.} \\[0.5em]

\textbf{Thought:} \\
\textit{The addition of 'carrot' creates a unique opening for novelty because carrots rarely coincide with couches, capsules, broth, illusion, or the ukiyo-e style itself, especially in a non-culinary, abstract context. To strategically maximize novelty, I must do more than simply depict a carrot in a bowl of broth. By treating carrots as modular architectural units, perhaps as the actual capsule shapes themselves or transformative couches, I can create something unexpected. I envision couches morphing into gigantic carrot forms, hollowed out as floating capsules within illusory swirling broth, blurring the line between furniture and vegetable. The women can be seated or reclining inside these carrot-capsule couch hybrids, surrounded by flowing mantle textiles patterned with root-vegetable motifs. To further highlight illusion, the carrots could be positioned so their tips and tops loop impossibly under and over the other elements, interacting with the heater patterns. I'll include 'woman', 'ukiyo\_e', 'carrot', 'couch', 'capsule', 'broth', 'illusion', and 'mantle', including both carrot and couch, as the fusion of their physical forms is unexpected, and omitting 'heater', 'mat', and 'position' to keep the scene visually coherent and maximize the focus on the newly serendipitous combinations.} \\[0.5em]

\textbf{Concepts Used:} \\
woman, ukiyo\_e, carrot, couch, capsule, broth, illusion, mantle
\end{tcolorbox}
\caption{Example response from the Prompt Compositor showing the adaptive combination of concepts to maximize novelty. The system demonstrates strategic reasoning by identifying unexpected but harmonious concept pairings (carrot-couch fusion) and explicitly not selecting other concepts to maintain visual coherence.}
\label{fig:prompt_compositor_example}
\end{figure}

\begin{figure}[H]
\centering
\begin{tcolorbox}[colback=gray!5!white, colframe=black!75!black, width=0.95\textwidth]
\small
\textbf{Generation:} 3 \\
\textbf{Name:} The Ukiyo-e Cyborg Huntress \\[0.5em]

\textbf{Prompt:} \\
\textit{A woman depicted in ukiyo-e style stands on a patch of snowy ground, framed by stylized pine branches overhead. She wears a pale blue kimono quilted at the shoulders, with brown hunter's boots. Her left arm is replaced by a segmented metallic robotic hunting prosthesis ending in an extended grip that holds a curved hunting knife with a serrated edge, the blade rendered in steel grey. Lying at her feet is a recently-caught hare, meticulously depicted with splayed hind legs and brown fur. The woman gazes downward, her black hair in a tight braid. The background has soft off-white sky shading typical of ukiyo-e prints, with faint mountains on the horizon. Shadows from the knife and machinery fall sharply across the snow. All elements are shown with technical accuracy, using firm black contours and flat color planes characteristic of ukiyo-e artwork.} \\[0.5em]

\textbf{Thought:} \\
\textit{With 'hunter' added to the pool alongside 'machinery' and 'knife', an especially rare and novel synthesis is possible when all are combined with the original concepts 'woman' and 'ukiyo\_e.' Traditionally, 'hunter' is not represented with advanced machinery, and the combination of a woman as a mechanized hunter in ukiyo-e style, using a knife as part of a hunting scenario, is highly unconventional. Previous novelty scores suggest that combining more concepts does not automatically guarantee higher novelty, but the key may be to create creative, non-trivial integration: not just stacking elements, but merging their functions. I will depict the woman as an explicit hunter character using machinery that assists in the hunt—perhaps a robotic hunting arm. This will create strong visual ties between 'woman', 'machinery', and 'hunter,' with the 'knife' as an integrated hunting tool. The aim is to move beyond juxtaposition, making the concepts inseparable in subject, action, and composition for maximum novelty.} \\[0.5em]

\textbf{Concepts Used:} \\
woman, ukiyo\_e, machinery, knife, hunter
\end{tcolorbox}
\caption{Example response from the Prompt Compositor showing strategic concept integration for novelty maximization. The system demonstrates sophisticated reasoning by identifying how to merge traditionally unrelated concepts (cyborg prosthetics, traditional hunting, ukiyo-e aesthetics) into a cohesive, unconventional synthesis rather than superficial juxtaposition.}
\label{fig:prompt_compositor_example_2}
\end{figure}

\subsection{CAS Validity Testing and Selection of $\beta$} \label{app-cas-validity}
To test the validity of CAS in selecting novel combinations of concepts, we compared the method, using various values of $\beta$ for Alien sampling, to a baseline method to generate novel concept combinations. The Baseline method generates sequences using the Concept Coherence Model but replaces the ranking with the Cultural Context Model and later selection of top-k with random sampling. 

In our experiment, we created 50 unique input sequences, each containing either 1 or 2 concepts. For each input sequence, we generated 150 output sequences at each temperature level in the Concept Coherence Model, ranging from 0.1 to 3.1 in increments of 0.3, for both CAS and Baseline method. From the generated sequences, we then selected the top-ranked sequence for each method based on its respective sampling strategy.

\paragraph{Assesing Text-based novelty}

Our methodology evaluates the novelty of concept combinations using two complementary measures.
\begin{itemize}
        \item \textbf{Novelty relative to artworks:} Let $A = \{A_1, A_2, \dots, A_m\}$ be the set of sets, where each $A_i \subset C$ represents the set of concepts in a single artwork within the Artwork dataset. Let $S \subset C$ be the set of concepts in the generated sequence. We define the novelty measure $N_{\text{art}}$ as: 
        \begin{equation}
        N_{\text{art}} = \min_{i \in \{1,\dots,m\}} |S \setminus A_i|
        \end{equation}
        This measure represents the minimum number of concepts in $S$ that do not appear together in any single artwork, when compared individually to each artwork in the dataset. A more intuitive way to interpret this measure is that it quantifies the number of distinct concepts in the generated artwork compared to the most similar artwork in the dataset.

        \item \textbf{Novelty relative to Artists:} Let $B = \{B_1, B_2, \dots, B_k\}$ be the set of sets, where each $B_j \subset C$ represents the set of all unique concepts used by artist $j$ in the Artwork dataset. We define the novelty measure $N_{\text{cog}}$ as:
        \begin{equation}
        N_{\text{cog}} = \min_{j \in \{1,\dots,k\}} |S \setminus B_j|
        \end{equation}
        This measure represents the minimum number of concepts in $S$ that do not appear together in any single artist’s cultural framework, when compared individually to each artist’s set of concepts. Thus, $N_{\text{cog}}$ quantifies the number of distinct concepts in the generated artwork compared to the most similar artist set. We frame this as a proxi of cultural availability.
    \end{itemize}

\begin{figure}[!ht]
    \centering
    \begin{subfigure}[b]{0.45\textwidth}
        \centering
        \includegraphics[width=\textwidth]{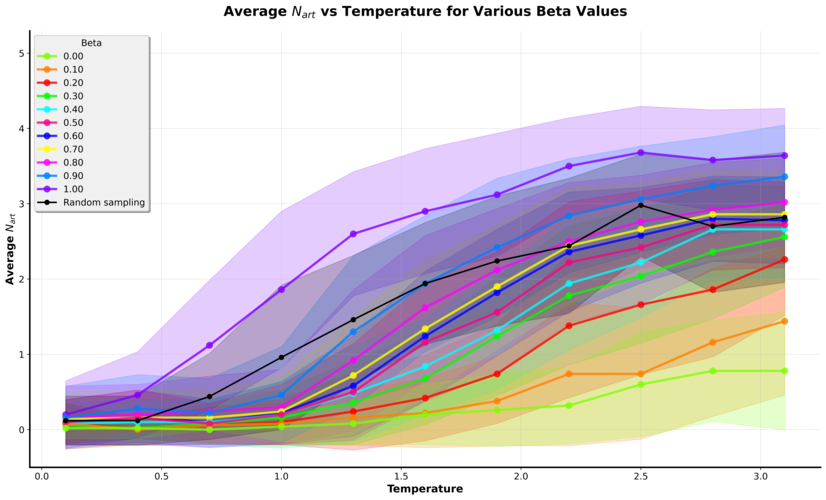}
        \caption{Average $N_{\text{art}}$ vs temperature for multiple values of $\beta$}
        \label{fig:art-dataset-beta}
    \end{subfigure}
    \hfill
    \begin{subfigure}[b]{0.45\textwidth}
        \centering
        \includegraphics[width=\textwidth]{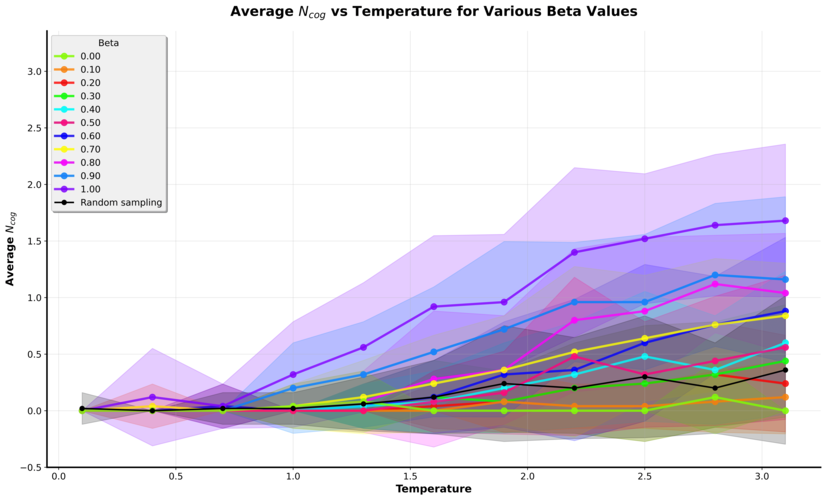}
        \caption{Average $N_{\text{cog}}$ vs temperature for multiple values of $\beta$}
        \label{fig:cog-dataset-beta}
    \end{subfigure}
    \caption{Comparison on the novelty of the generated sequences with respect to the artworks and with respect to artists full conceptual coverage for multiple values of $\beta$}
    \label{fig:comparison-dataset-beta}
\end{figure}
    
Our findings reveal that while increasing the Concept Coherence model's temperature can generate novel combinations absent from the dataset at artwork-level (Figure \ref{fig:art-dataset-beta}), it does not reliably produce novel combinations at artist-level. CAS method, through its explicit search for artist-level novelty, demonstrates a consistently higher likelihood of generating such combinations (Figure \ref{fig:cog-dataset-beta}). Empirically, we show that generating unseen combinations in artworks when increasing the temperature is relatively straightforward, with 85\% of the combinations containing at least one new concept when surpassing temperature 1. However, this phenomenon does not extend to artist-level combinations: at temperature 3, more than half of the combinations have $N_{\text{cog}}= 0$, indicating no artist-level novel concepts given the rest of the generated sequence. This shows that identifying concepts unused by any artist, given that they have used a set of other concepts, is fundamentally more difficult and requires explicit search strategies within the concept space.

\begin{figure}[!htbp]
    \centering
    % Temperature 0.1 PArt
    \begin{subfigure}[b]{0.48\textwidth}
        \centering
        \includegraphics[width=\textwidth]{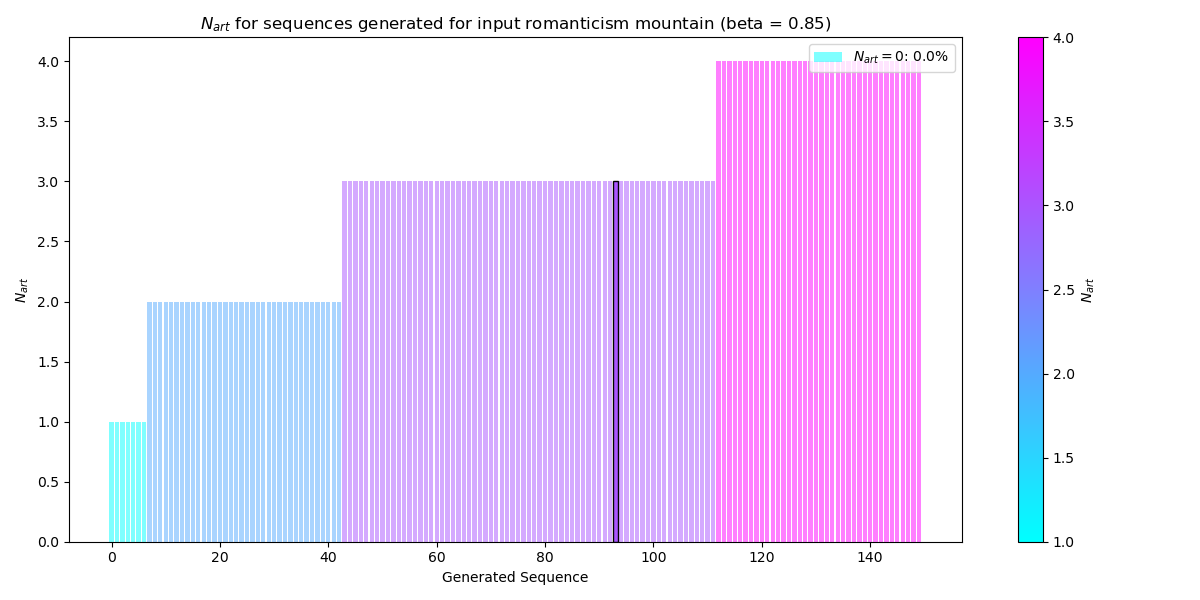}
        \caption{$N_{\text{art}}$ with $\beta = 0.85$.}
    \end{subfigure}
    \hspace{0.5mm} % Small space between images
    % Temperature 0.1 PCog
    \begin{subfigure}[b]{0.48\textwidth}
        \centering
        \includegraphics[width=\textwidth]{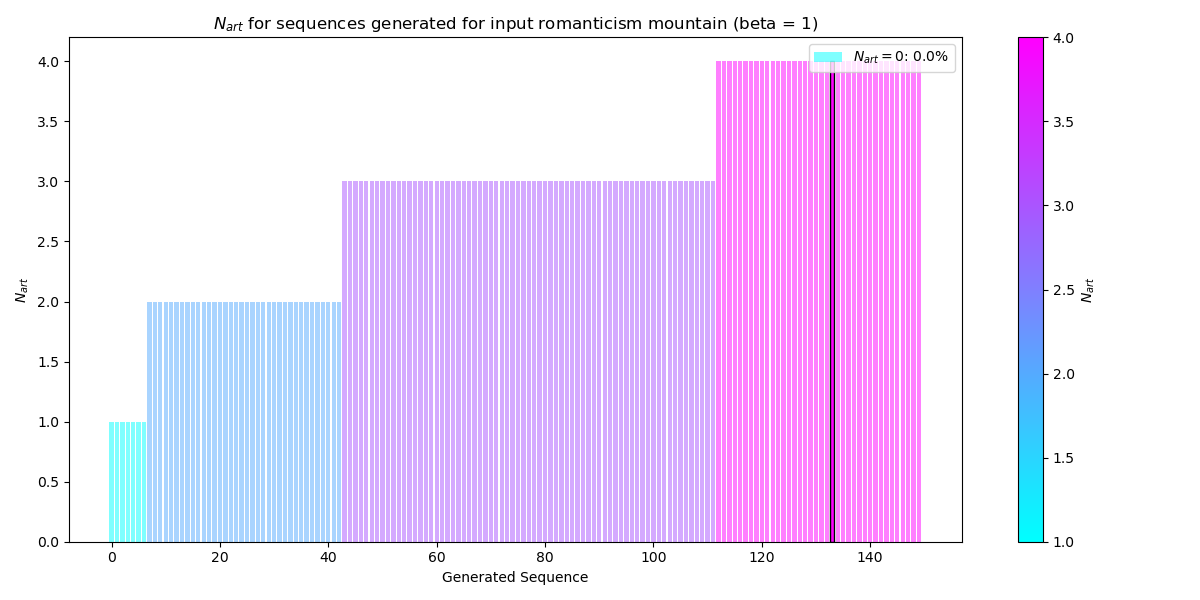}
        \caption{$N_{\text{art}}$ with $\beta = 1$.}
    \end{subfigure}
    % Temperature 0.7 PArt
    \begin{subfigure}[b]{0.48\textwidth}
        \centering
        \includegraphics[width=\textwidth]{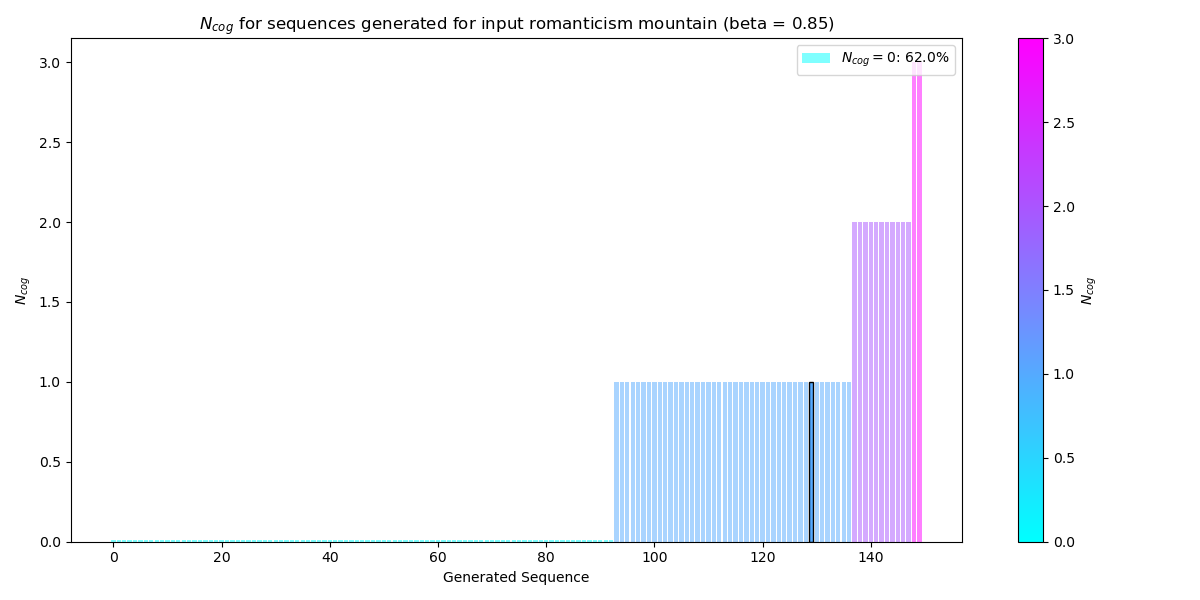}
        \caption{$N_{\text{cog}}$ with $\beta = 0.85$.}
    \end{subfigure}
    \hspace{0.5mm} % Small space between images
    % Temperature 0.7 PCog
    \begin{subfigure}[b]{0.48\textwidth}
        \centering
        \includegraphics[width=\textwidth]{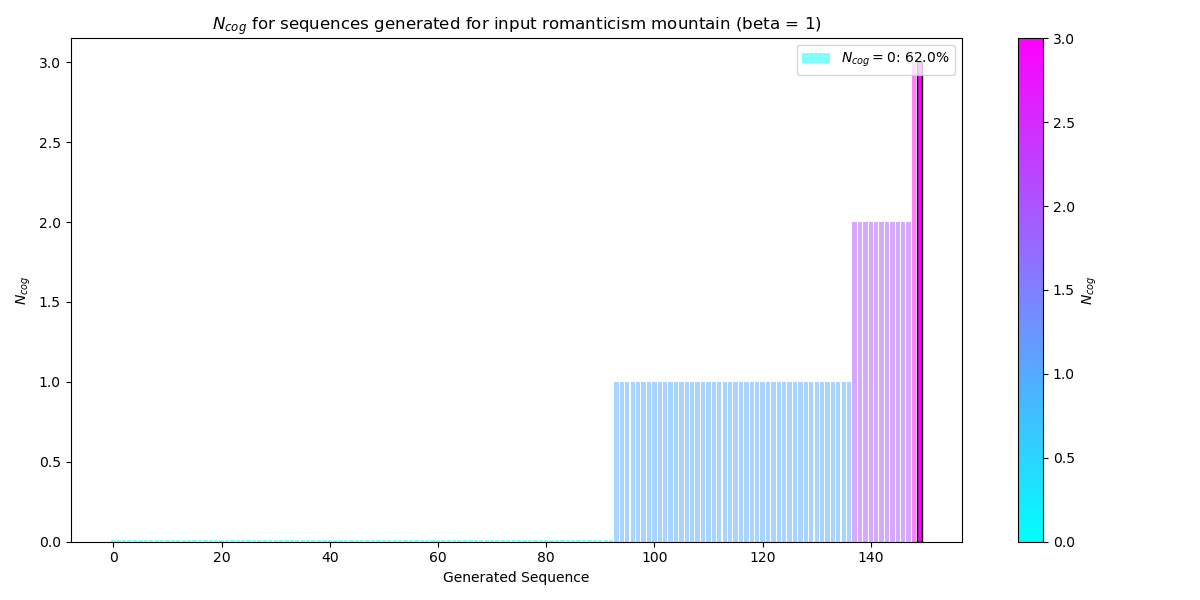}
        \caption{$N_{\text{cog}}$ with $\beta = 1$.}
    \end{subfigure} 
    \caption{Comparison between selected sequences with CAS with $\beta = 0.85$ and $\beta = 1$ for input "Romanticism Mountain" and temperature 2.5. The sequences are sorted by $N_{\text{art}}$ and $N_{\text{cog}}$, depending on the measure plotted. The selected sequence is highlighted in black. The proportion of not cultural unavailable ($N_{\text{cog}} = 0$) sequences is in the right corner.}
    \label{fig:comparison-beta-n}
\end{figure}

\paragraph{Artwork-Level Novelty is Easy, Artist-Level Novelty is Hard}
We plotted \(N_{\text{art}}\) and \(N_{\text{cog}}\) for each sequence generated by the Concept Coherence Model across various temperature levels, highlighting the selected sequence using CAS with $\beta = 0.85$ (Figure \ref{fig:comparison-beta-n}). The results clearly show that, even at high temperatures, finding novel artist-level sequences is much harder than finding artwork-level novel ones. For example, at a temperature of 2.5, all generated sequences have \(N_{\text{art}} \geq 1\), with many sequences showing \(N_{\text{art}} \geq 3\). However, 62.0\% of these sequences are not culturally unavailable (\(N_{\text{cog}} = 0\)), with a notable difference in density. This highlights the need for a method that explicitly detects and selects these combinations, as merely temperature scaling has proven to be an inefficient approach.

Finally, the impact of \(\beta\) on sequence selection is evident in Figure \ref{fig:comparison-beta-n}. For the same input and temperature level, \(\beta = 1\) typically selects the sequence with the highest \(N_{\text{cog}}\). In contrast, \(\beta = 0.85\) occasionally selects sequences with lower \(N_{\text{cog}}\), reflecting a trade-off between maximizing cultural unavailability and favoring sequences that remain plausible within the context of existing artworks, a balance well-suited to generating original yet coherent artwork ideas in an open-ended way.

\clearpage

\subsection{Experimental Details}

The following experiments were approved by the Institutional Review Board at Max-Planck Institute for Human Development (C2025-08).

\subsubsection{Human Sampling}
\label{app:experiment-human}

\begin{figure}
    \centering
    \includegraphics[width=0.8\linewidth]{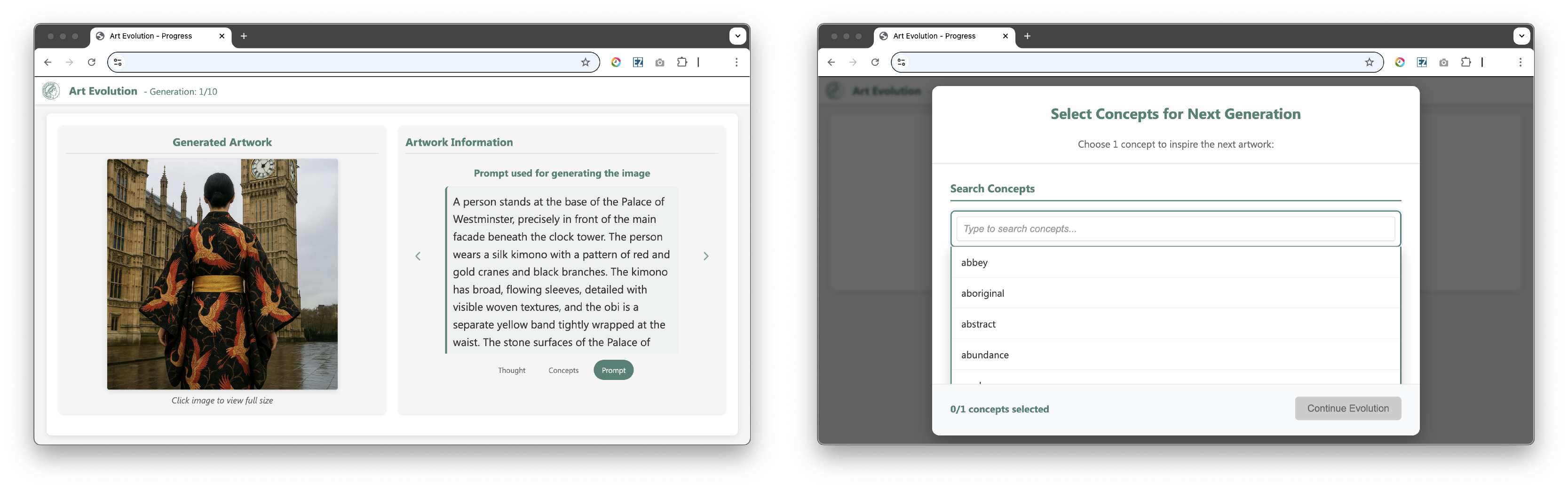}
    \caption{A web interface to interact with Open-ended Art Agent. (Left) Based on the combination of the selected concepts, the art students were able to see the generation results. (Right) The art students iteratively explored the combination of art concepts.}
    \label{fig:interface}
\end{figure}

To prepare a reasonable baseline, we involved humans who possess expert knowledge about art concepts as a source of inspiration. Specifically, we recruited 16 art students from the BA, MA, and PhD levels through a mailing list in two art schools: the Baltic Film, Media and Arts School of Tallinn University (Estonia) and the Academy of Media Arts Cologne (Germany). They interacted with the Open-ended Art Agent through a Web interface (Figure~\ref{fig:interface}), in which they were first asked to decide the seed concept. They then explored the combination of new concepts through the iterative visualization from the Prompt Compositor and Image Generator. Here, in each iteration, they were presented with a potential combination proposed by the suggestion from either GPT-4o sampling or CAS as an example. We confirmed their adoption rate of the suggestion was approximately 15~\%, where they played the role of curator.

\subsubsection{Human Evaluation}
\label{app:experiment-evaluation}

To evaluate the reception of the generated images by the general audience, we recruited two independent pools of 50 human raters, each for originality and harmony, on Prolific. Each rater was presented with two images, generated from the same seed concepts but with different samplers. They answered which of the two images is better in terms of originality and harmony, based on the following definitions (Figures~\ref{fig:originality} and \ref{fig:harmony}, respectively).

\begin{figure}[H]
\centering
\begin{mdframed}[linewidth=1pt, linecolor=gray!50, backgroundcolor=gray!5]
\small
\textbf{What We Mean by ``Original''?}

When evaluating originality, consider the following questions:
\begin{itemize}
    \item Is the combination of concepts in the image original and interesting?
    \item Have I ever seen or imagined a painting like this before? And at the same time, is the painting itself, or the source of its originality, interesting or valuable?
    \item Does the visual presentation feel original or unique?
\end{itemize}
There are no right or wrong answers, we're interested in your personal judgment of originality.
\end{mdframed}
\caption{Definition of the originality presented to the human raters}
\label{fig:originality}
\end{figure}

\begin{figure}[H]
\centering
\begin{mdframed}[linewidth=1pt, linecolor=gray!50, backgroundcolor=gray!5]
\small
\textbf{What We Mean by ``Harmony''?}

When evaluating harmony, consider the following questions:
\begin{itemize}
    \item Do the concepts/elements make sense together artistically or aesthetically, regardless of whether they're typically combined?
    \item Do the elements in the image harmonize with each other visually?
    \item Do the concepts complement each other in a way that feels meaningful rather than forced?
\end{itemize}
There are no right or wrong answers, we're interested in your personal judgment of harmony.
\end{mdframed}
\caption{Definition of the harmony presented to the human raters}
\label{fig:harmony}
\end{figure}

\subsubsection{GPT concept repetition details} \label{app-repetition}

For Free GPT repetition analysis, we empirically evaluated similarity thresholds of 0.9, 0.87, and 0.85 to optimize the detection of conceptually equivalent ideas in the simulations. While higher thresholds (0.9, 0.87) provided greater precision by filtering out weakly related concepts, they proved overly restrictive, failing to capture semantically meaningful relationships that humans would clearly consider equivalent. At the optimal threshold of 0.85, our system successfully identified genuine conceptual repetitions such as 'bioluminescent forest' and 'bioluminescence' (similarity: 0.860) or 'ancient rituals' and 'ancient civilizations' (similarity: 0.858), all pairs that represent the same core conceptual domain despite lexical variation. Conversely, the threshold effectively filtered out conceptually distinct pairs such as 'ancient rituals' vs 'cyberspace' (similarity: 0.786) or 'time travel' vs 'cyberpunk' (similarity: 0.751). Thus, we found that this threshold strikes the optimal balance between recall for true semantic equivalences and precision against false conceptual matches, aligning with CLIP's learned representation space where genuine conceptual similarity clusters around the 0.85-0.86 range.

Using this optimized threshold of 0.85, our analysis revealed significant repetition patterns across different generation methods. As shown in the \textit{Experiments and results} section, GPT and Free GPT exhibited the highest levels of repetition among the inspiration methods, with average repetition rates of 59\% and 74.3\%, respectively. In other words, on average more than half of the generated concepts have already appeared in other simulations with different initial concepts, with some simulations reaching complete (100\%) overlap. This high repetition rate suggests that current LLM-based inspiration systems may suffer from limited conceptual diversity, potentially constraining creative exploration.

\begin{figure}[ht]
    \centering
    \begin{subfigure}[b]{0.45\textwidth}
        \centering
        \includegraphics[width=\textwidth]{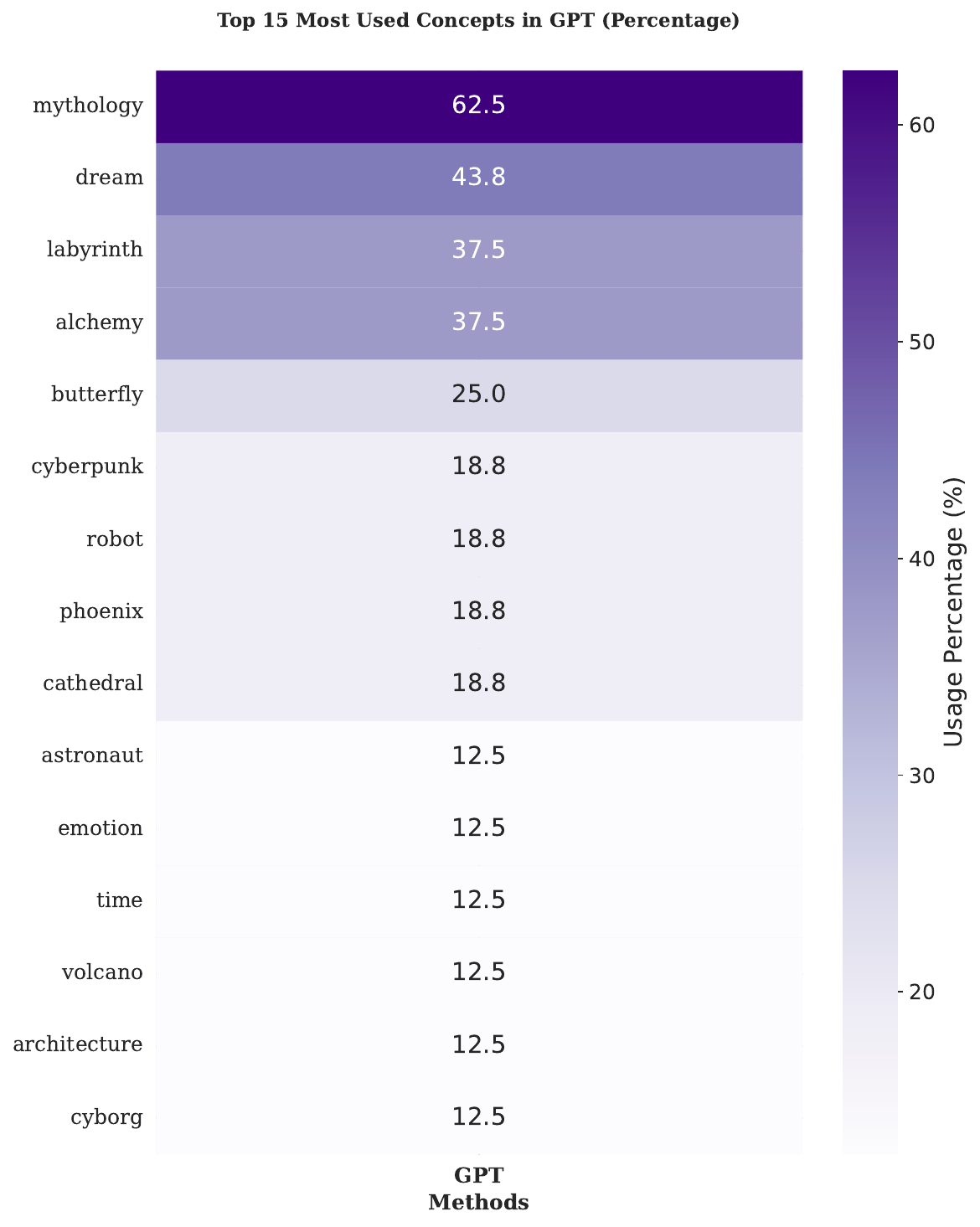}
        \caption{}
        \label{fig:gpt-repetition}
    \end{subfigure}
    \begin{subfigure}[b]{0.47\textwidth}
        \centering
        \includegraphics[width=\textwidth]{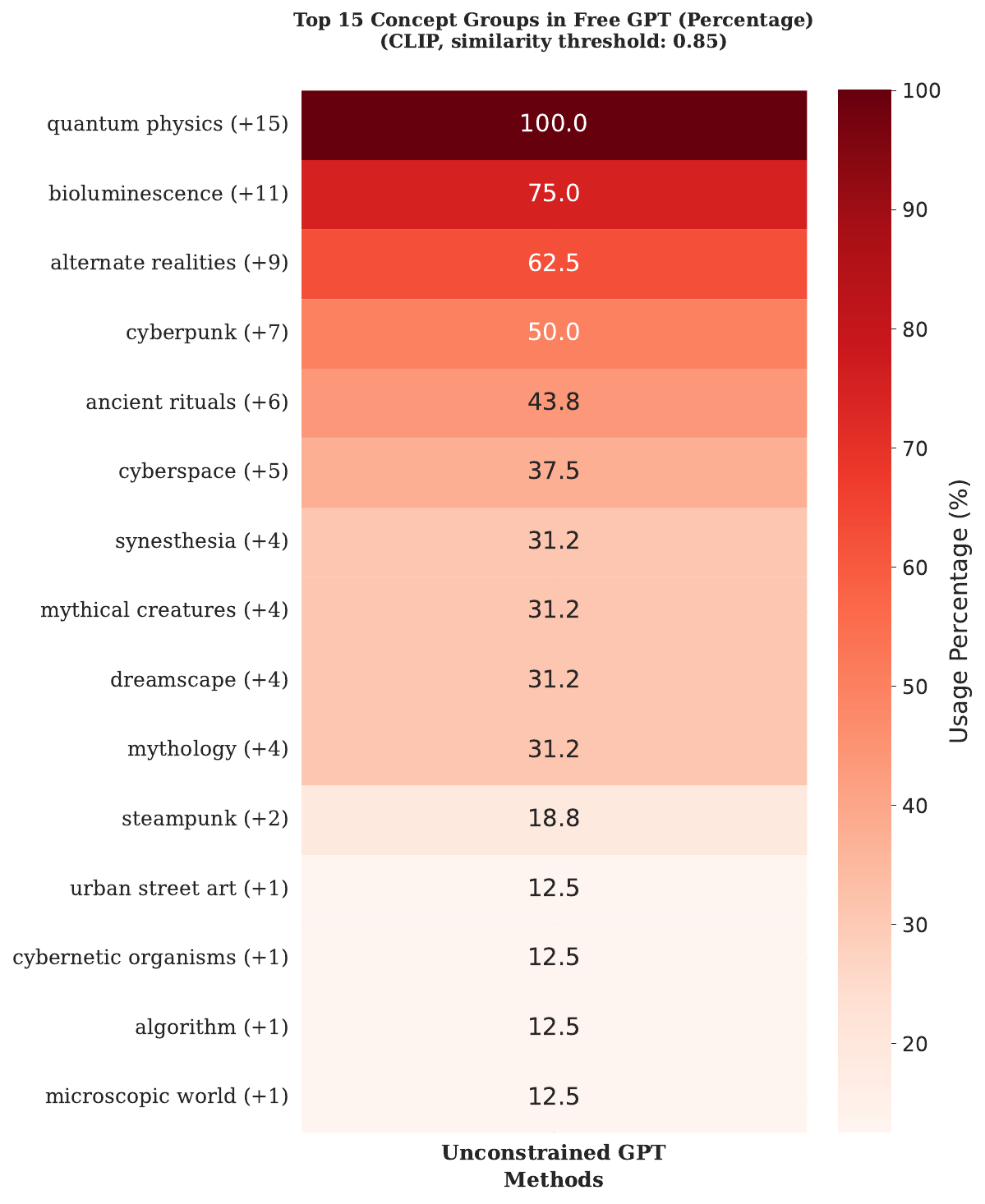}
        \caption{}
        \label{fig:free-gpt-repetition}
    \end{subfigure}
    \caption{(Left) Top 15 most frequently repeated concepts across runs for the GPT inspiration method. (Right) Top 15 most frequently repeated concepts across runs for Free GPT. The notation (+N) next to a concept indicates the number of additional distinct concepts that also met the repetition threshold in relation to that concept.}
    \label{fig:all-gpt-repetition}
\end{figure}

Figure \ref{fig:all-gpt-repetition} illustrates this tendency: in both versions, GPT, despite being explicitly prompted for originality, displays a bias toward a recurring set of themes. These often include bioluminescence, futuristic and science-fiction aesthetics (e.g., quantum, cyberpunk, steampunk, cyberspace), as well as mythology and dreamlike scenarios.

\clearpage
\subsubsection{Long-Range Exploration details} \label{app-long-range}

\textbf{Return Rate Computation}: For each generation $t > 1$, we calculate the minimum Euclidean distance between the current embedding and all previous embeddings in the trajectory. A "return" occurs when this minimum distance falls below a threshold set to the median step distance for that specific inspiration method (making the metric adaptive to each method's natural exploration scale). The return rate is then computed as the fraction of all generations (excluding the first) that qualify as returns, providing a measure of how frequently an agent revisits semantically similar regions relative to its typical exploration granularity.

\textbf{Saturation Generation Computation}: We first compute the exploration radius for each generation as the Euclidean distance from the starting embedding to the current generation's embedding. The saturation generation is defined as the first generation where this radius reaches or exceeds 95\% of the maximum radius achieved throughout the entire trajectory. This metric captures how quickly an agent reaches its peak exploration distance, distinguishing between methods that rapidly expand to their boundaries versus those that gradually explore outward over many generations. Fast saturation typically correlates with high return rates, as methods that quickly reach their exploration boundaries subsequently engage in local search behavior, repeatedly revisiting familiar semantic regions within their established trajectory.

\begin{figure}[h]
    \centering
    \begin{subfigure}[b]{0.48\textwidth}
        \centering
        \includegraphics[width=\textwidth]{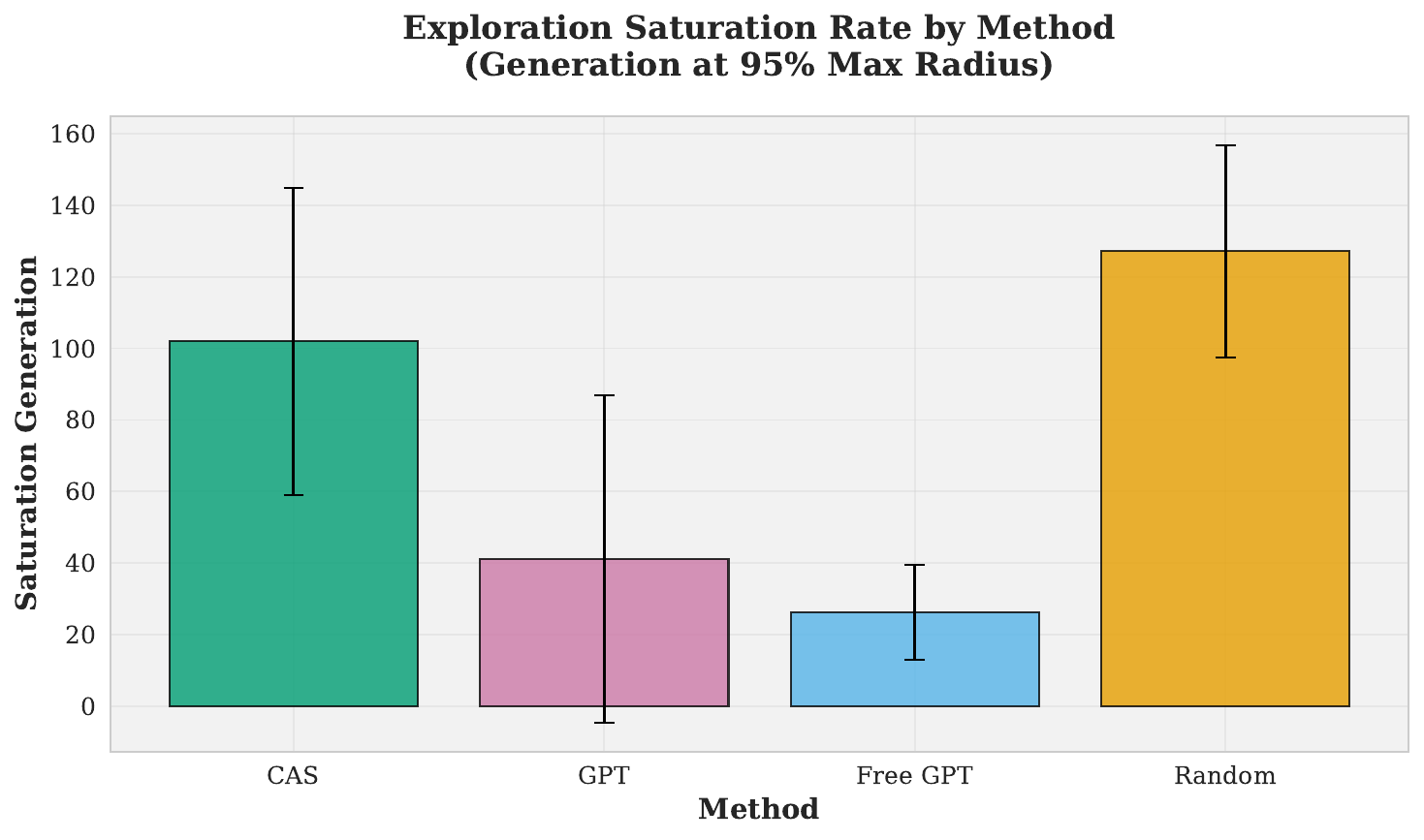}
        \caption{}
        \label{fig:return rate}
    \end{subfigure}
    \begin{subfigure}[b]{0.48\textwidth}
        \centering
        \includegraphics[width=\textwidth]{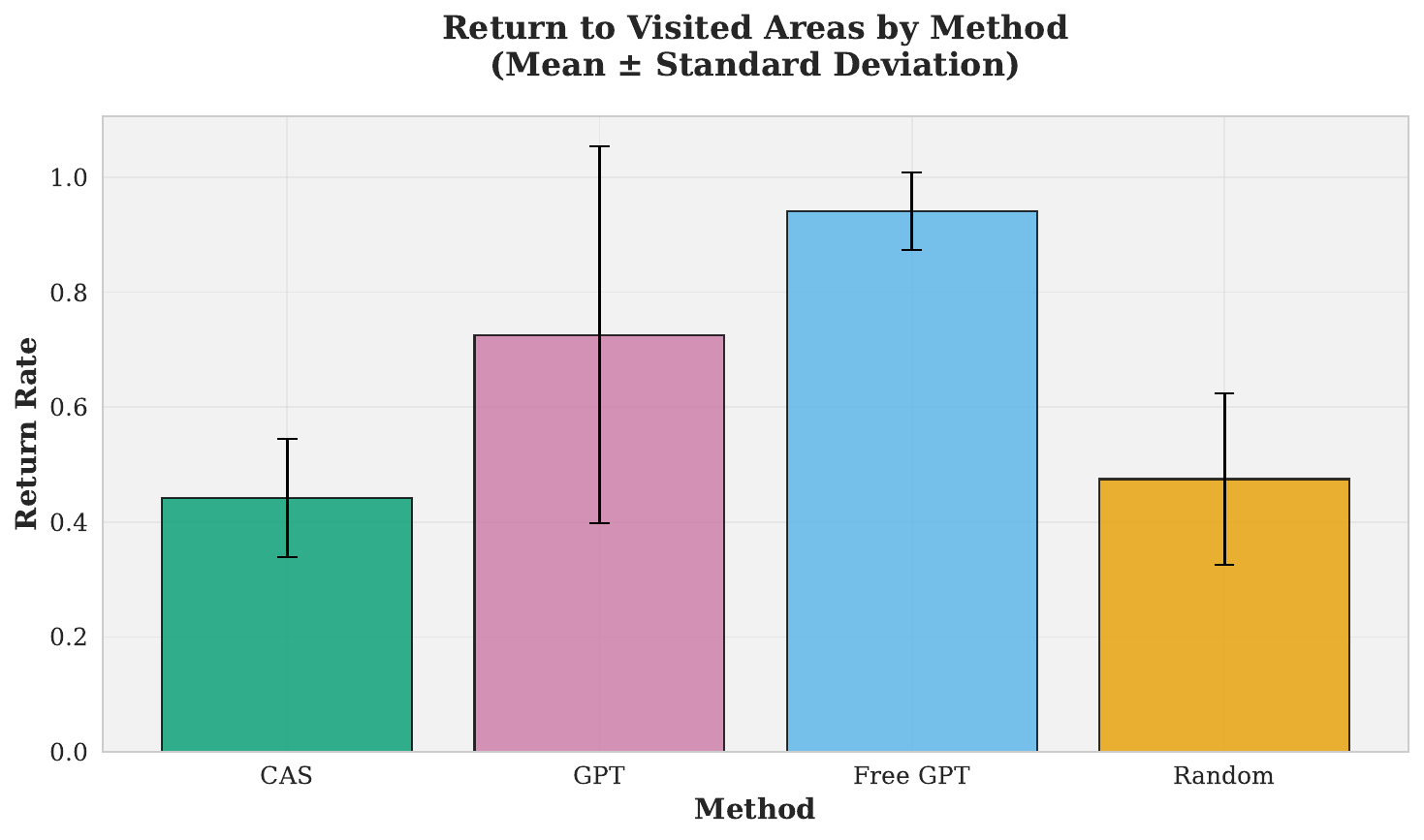}
        \caption{}
        \label{fig:saturation}
    \end{subfigure}
    \caption{\textbf{Exploration behavior across agent types}: Saturation analysis (a) shows distinct strategies: CAS agents sustain exploration with late saturation (100 generations), Free GPT saturates fastest (25) then focuses on local search, Standard GPT is intermediate (~40), and Random agents saturate slowest (130) due to random wandering. Return rate analysis (b) aligns with these patterns: Free GPT revisits most (0.95), CAS and Random are lower (0.45–0.48) due to more diverse conceptual exploration, while  GPT is moderate (0.72). Together, the metrics show fast saturation coincides with high revisitation, reflecting local search after boundaries are reached.}
    \label{fig:long-range-appendix}
\end{figure}

\begin{figure}[t]
    \centering
    \includegraphics[width=0.9\textwidth]{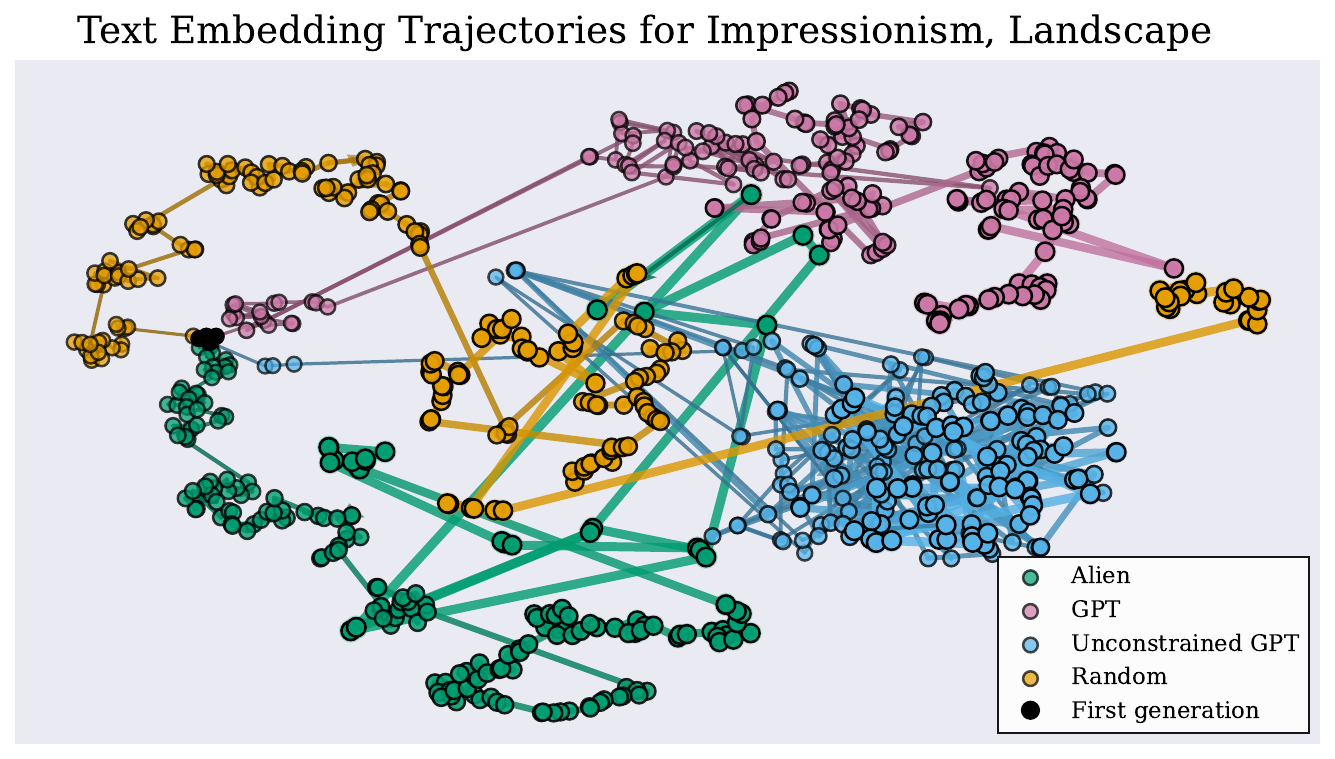}
    \caption{Example of 200 generations per method for starting concepts \textit{Impressionism, Landscape}, showing trajectories in 2D t-SNE space by embedding each artwork’s prompt. Larger points and wider edges indicate later generations.}
    \label{fig:impressionism-landscape-trajectory}
\end{figure}

\end{document}